\documentclass[10pt,twocolumn,letterpaper]{article}

\usepackage[pagenumbers]{cvpr} 

\usepackage{graphicx}
\usepackage{amsmath}
\usepackage{amssymb}
\usepackage{booktabs}

\usepackage{comment}
\usepackage{color}
\usepackage{overpic}
\usepackage{multirow}
\usepackage{tabularx}
\usepackage{adjustbox}
\usepackage{times}
\usepackage{epsfig}
\usepackage{float}

\usepackage[pagebackref,breaklinks,colorlinks]{hyperref}

\usepackage[capitalize]{cleveref}
\crefname{section}{Sec.}{Secs.}
\Crefname{section}{Section}{Sections}
\Crefname{table}{Table}{Tables}
\crefname{table}{Tab.}{Tabs.}

\def\cvprPaperID{3788} 
\def\confName{CVPR}
\def\confYear{2022}

\begin{document}

\title{A Differentiable Two-stage Alignment Scheme for Burst Image \\ Reconstruction with Large Shift}

\author{Shi Guo\textsuperscript{1} \quad Xi Yang\textsuperscript{1} \quad Jianqi Ma\textsuperscript{1} \quad Gaofeng Ren\textsuperscript{2} \quad Lei Zhang\textsuperscript{1} \\
\textsuperscript{1}The Hong Kong Polytechnic University; \textsuperscript{2}DAMO Academy, Alibaba Group\\
{\tt\small \{shiguo.guo,xxxxi.yang,jianqi.ma\}@connect.polyu.hk,} \\
{\tt\small gaof.ren@gmail.com,  cslzhang@comp.polyu.edu.hk}}
\maketitle

\begin{abstract}
   Denoising and demosaicking are two essential steps to reconstruct a clean full-color image from the raw data. Recently, joint denoising and demosaicking (JDD) for burst images, namely JDD-B, has attracted much attention by using multiple raw images captured in a short time to reconstruct a single high-quality image. One key challenge of JDD-B lies in the robust alignment of image frames. State-of-the-art alignment methods in feature domain cannot effectively utilize the temporal information of burst images, where large shifts commonly exist due to camera and object motion. In addition, the higher resolution (e.g., 4K) of modern imaging devices results in larger displacement between frames. To address these challenges, we design a differentiable two-stage alignment scheme sequentially in patch and pixel level for effective JDD-B. The input burst images are firstly aligned in the patch level by using a differentiable progressive block matching method, which can estimate the offset between distant frames with small computational cost. Then we perform implicit pixel-wise alignment in full-resolution feature domain to refine the alignment results. The two stages are jointly trained in an end-to-end manner. Extensive experiments demonstrate the significant improvement of our method over existing JDD-B methods. Codes are available at \url{https://github.com/GuoShi28/2StageAlign}.
\end{abstract}

\section{Introduction}
\label{sec:intro}
Color demosaicking and denoising are two essential steps in digital camera imaging pipeline to reconstruct a high quality full-color image from the sensor raw data. Color demosaicking~\cite{ehret2019study,yang2019efficient,yan2019cross,liu2020new,jin2020review} recovers the missing color components from the color-filter-array (CFA) data collected by the single-chip CCD/CMOS sensor, while denoising~\cite{luisier2010image,khademi2021self,zhang2017beyond,zhang2018ffdnet} removes the noise in image data caused by the photon arrival statistics and the imprecision in readout circuitry. Since the two tasks are actually correlated and can be performed jointly, many joint denoising and demosaicking (JDD) algorithms have been developed~\cite{condat2012joint,heide2014flexisp,gharbi2016deep,henz2018deep,kokkinos2019iterative,liu2020joint}, which are however focused on single image restoration. With the prevalent use of smartphone cameras~\cite{CIPA_2018}, it becomes crucial to restore images from data with low signal-to-noise ratio due to the small sensor and lens of smartphone cameras. To this end, performing JDD with burst images (JDD-B) has become increasingly popular and important in recent years~\cite{guo2021joint}.

Burst image processing refers to shooting a sequence of low quality frames in a short time and computationally fusing them to produce a higher-quality photograph~\cite{aittala2018burst,wronski2019handheld}. Compared with single image restoration, the key challenge of burst image restoration lies in the compensation for shift between frames. Previous studies often perform frame alignment by estimating optical flow~\cite{ehret2019model,xue2019video} and applying spatially variant kernels~\cite{mildenhall2018burst,xu2019learning,marinvc2019multi,xia2019basis} in the image domain. However, affected by noise, accurately estimating optical flow and kernels is difficult. Recently, implicit alignment in the feature domain has achieved state-of-the-art performance on video super-resolution~\cite{wang2019edvr,chan2020basicvsr}, video denoising~\cite{yue2020supervised}, as well as JDD-B~\cite{guo2021joint}. However, it is found that feature alignment is not very effective when handling image sequences with large shift. For example, for the pyramid feature alignment module in \cite{wang2019edvr}, the receptive field of offset estimation is about 28 (3 conv layers with $3\times 3$ kernels on $1/4$ scale), \emph{i.e.}, 14 pixels along one direction. Such a search range is too small for burst images with large shift. Whereas, large shifts commonly exist in videos due to the camera and object motion. On the other hand, modern image/video recorders can easily capture videos of 4K ($4096\times  2160$) or UHD ($3840\times 2160$) resolution. As a result, the pixel displacements between frames further increase. Even some small motion of foreground objects in 4K video can cause a large value of shift. 

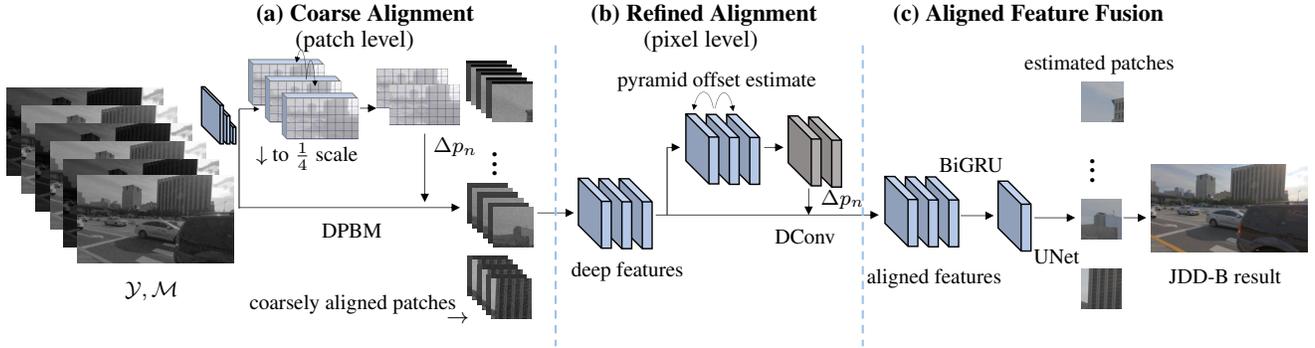
\begin{figure*}[!t]
\centering
\vspace{0.6cm}
\begin{overpic}[width=1.0\textwidth]{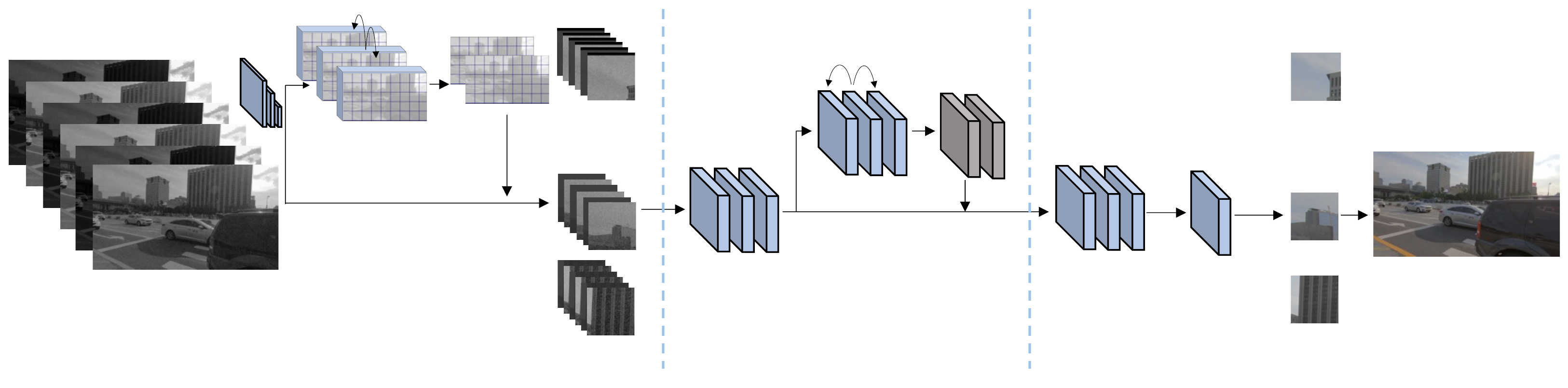}
	\put(9.5,5){\color{black}{\footnotesize $\mathcal{Y}, \mathcal{M}$}}
	\put(19.5,26){\color{black}{\small \textbf{(a) Coarse Alignment}}}
	\put(22.5,24){\color{black}{\small (patch level)}}
	\put(37.5,14){\color{black}{\Large \rotatebox{90}{...}}}
	\put(83.0,13.5){\color{black}{\Large \rotatebox{90}{...}}}
	\put(24.5,9.5){\color{black}{\footnotesize DPBM}}
	\put(33.0,16.0){\color{black}{\footnotesize $\Delta p_n$}}
	\put(19.5,15.3){\color{black}{\footnotesize $\downarrow$ to $\frac{1}{4}$ scale}}
	\put(19,4){\color{black}{\footnotesize coarsely aligned patches}}
	\put(34, 3){\color{black}{\footnotesize $\rightarrow$}}
	
	\put(43.5,6.5){\color{black}{\footnotesize deep features}}
	\put(47,21){\color{black}{\footnotesize pyramid offset estimate}}
	\put(62.5,12){\color{black}{\footnotesize $\Delta p_n$}}
	\put(59,9.0){\color{black}{\footnotesize DConv}}
	 
	\put(45,26){\color{black}{\small \textbf{(b) Refined Alignment}}}
	\put(49,24){\color{black}{\small (pixel level)}}
	
	\put(66,6){\color{black}{\footnotesize aligned features}}
	\put(71.5,14.5){\color{black}{\footnotesize BiGRU}}
	\put(78.7,7.5){\color{black}{\footnotesize UNet}}
	\put(78,22.3){\color{black}{\footnotesize estimated patches }}
	\put(89,6){\color{black}{\footnotesize JDD-B result }}
	\put(68,26){\color{black}{\small \textbf{(c) Aligned Feature Fusion}}}
	
\end{overpic}\vspace{-0.2cm}
\caption{Illustration of our network with differentiable two-stage alignment for JDD-B.}
\label{figOverall}
\vspace{-0.2cm}
\end{figure*}

However, precise pixel-wise alignment on full size images with large motion is very difficult and expensive. With a naive implementation, for each pixel, we assume the offset estimation with receptive field $D\times D$ costs $F\times D^2$ multiply-adds, where $F$ is determined by different network structures. For input with size $H\times W$, the offset estimation costs $F\times D^2\times H\times W$ multiply-adds. One straightforward solution to handle large shift is to increase $D$ by using 3D conv layers~\cite{liu2021large} or calculating all-range correlation volume~\cite{li2021arvo}. However, these solutions will significantly increase the computational cost and they are not efficient for images with large size. 

To address these problems, we design a differentiable two-stage alignment framework, which divides the difficult large shift alignment problem into two relatively easier alignment sub-problems, \emph{i.e.,} coarse alignment (CA) and refined alignment (RA). The CA module aims to compensate large shift roughly using small calculating resources. Instead of using pixel-wise alignment, the CA module performs alignment in patch level with $F\times D^2\times (H/k)\times (W/k)$ multiply-adds, where $k$ is the patch size. Then the RA module is used to pixel-wise align frames based on the results of CA module with smaller receptive field $D_s$. Such two-stage framework uses $F\times (D^2/k^2+D_s^2)\times H\times W$ multiply-adds in total, which is significantly smaller than directly aligning images using one-stage module, especially when $D$ is large. Specifically, we utilize block matching (BM) based method as CA module in image domain. 

To overcome the non-differentiability caused by BM and reduce the computational cost, we propose a differentiable progressive block matching (DPBM) method. We further propose a corresponding loss function for DPBM to stablize the training process. For the RA module, we perform refined pixel-wise alignment implicitly in the feature domain using deformable convolution for accurate image restoration. The two stages of alignments are complementary and they can enhance each other when jointly used in our learning framework. Compared with state-of-the-art (SOTA) method GCP-Net, our two-stage framework has fewer learnable parameters and similar running time, but achieves great improvement on images with large shift. The major contributions of this work are shown as follows:
\begin{itemize}
\setlength{\topsep}{0pt}
\setlength{\itemsep}{0pt}
\setlength{\parsep}{0pt}
\setlength{\parskip}{0pt}

\item To efficiently handle images with large shift, we propose a differentiable two-stage alignment framework, which divides the difficult large shift alignment problem into two easier sub-problems.
\item We propose a differentiable progressive block matching method in CA module to reduce computational cost and ensure that our two-stage framework can be end-to-end trained. 
\end{itemize}

Our experiments on synthetic and real-world burst image datasets clearly show that our two-stage alignment method achieves impressive improvement over existing methods.

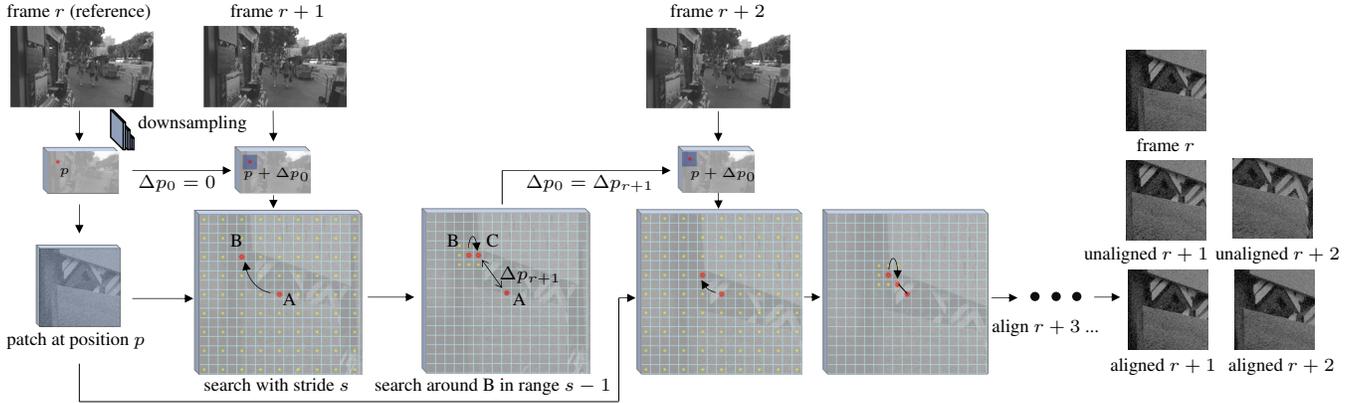
\begin{figure*}[!t]
\centering
\begin{overpic}[width=1.0\textwidth]{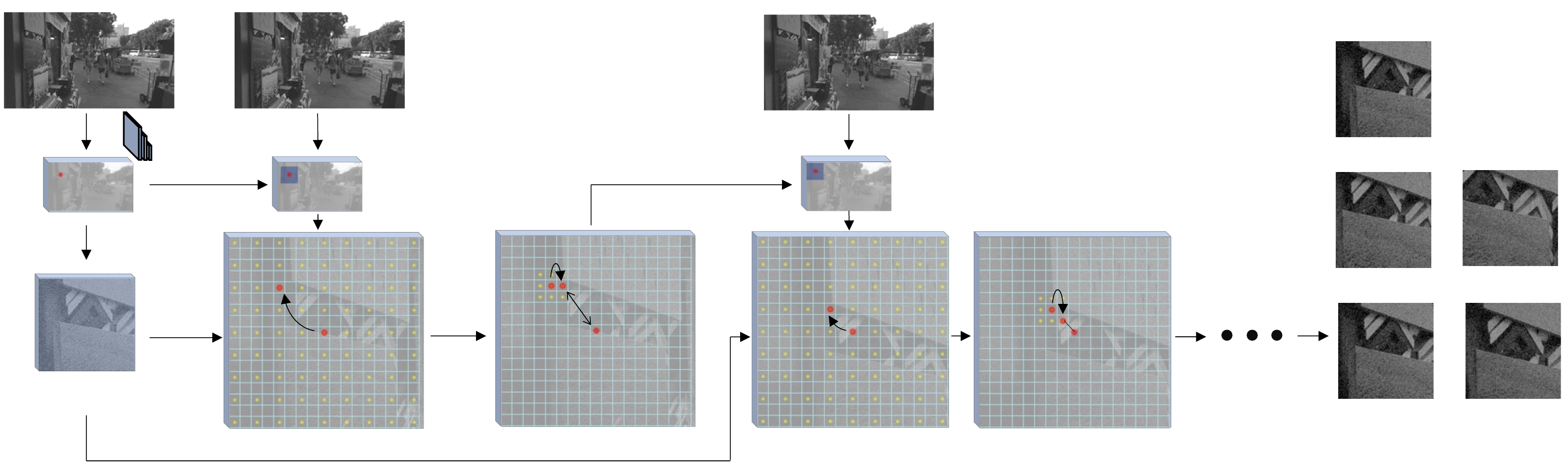} 
    \put(0,30){\color{black}{\scriptsize  frame $r$ (reference)}}
    \put(17.0,30){\color{black}{\scriptsize  frame $r+1$}}
    \put(50.5,30){\color{black}{\scriptsize  frame $r+2$}}
    
    \put(10,21.5){\color{black}{\scriptsize  downsampling}}
    \put(4,18){\color{black}{\tiny  $p$}}
    \put(10,17.0){\color{black}{\scriptsize  $\Delta p_0=0$}}
    \put(18.0,17.8){\color{black}{\tiny  $p + \Delta p_0$}}
    
    \put(0,5){\color{black}{\scriptsize patch at position  $p$}}
    \put(21,8){\color{black}{\scriptsize A}}
    \put(17,12.5){\color{black}{\scriptsize B}}
    \put(15,1.5){\color{black}{\scriptsize search with stride $s$}}
    \put(28,1.5){\color{black}{\scriptsize search around B in range $s-1$}}
    \put(38.5,8){\color{black}{\scriptsize A}}
    \put(33.5,12.5){\color{black}{\scriptsize B}}
    \put(36.5,12.5){\color{black}{\scriptsize C}}
    \put(37.5,10.0){\color{black}{\scriptsize $\Delta p_{r+1}$}}
    \put(39.5,17.0){\color{black}{\scriptsize  $\Delta p_0=\Delta p_{r+1}$}}
    \put(52,17.8){\color{black}{\tiny  $p + \Delta p_0$}}
    
    \put(75,6){\color{black}{\scriptsize  align $r+3$ ...}}
    \put(86,19.8){\color{black}{\scriptsize  frame $r$}}
    \put(82,11.5){\color{black}{\scriptsize  unaligned $r+1$}}
    \put(92,11.5){\color{black}{\scriptsize  unaligned $r+2$}}
    
    \put(84,3){\color{black}{\scriptsize  aligned $r+1$}}
    \put(93,3){\color{black}{\scriptsize  aligned $r+2$}}
    
\end{overpic}
\caption{Illustration of the coarse alignment process with differentiable progressive block matching (DPBM). Without loss of generality, we show the DPBM of a patch centered at position $p$. ``A" is the center of search region centered at $p+\Delta p_0$, while $\Delta p_0$ is 0 for target frame $r+1$ and $\Delta p_{r+n}$ for target frame $r+n+1$. To reduce computational cost, we first perform matching with stride $s$ using proposed differentiable block matching, resulting in the matched center position ``B". Then we perform searching with stride 1 in the neighborhood of ``B", resulting in a more accurate matched position ``C". $\Delta p_{r+1}$ is the displacement between ``C" and ``A" for frame $r + 1$.}
\label{figPBM}
\vspace{-0.4cm}
\end{figure*}

\section{Related Work}
\label{sec:ref}
\subsection{Joint Denoising and Demosaicking}
Denoising and demosaicking are two fundamental and correlated tasks in camera image signal processing (ISP) pipeline with a single chip of CMOS/CCD sensor~\cite{ehret2019study,yang2019efficient,yan2019cross,liu2020new,jin2020review}. Considering that performing denoising and demosaicking separately may accumulate errors for image restoration, JDD has been widely studied in recent years~\cite{gharbi2016deep,qian2019trinity,henz2018deep,kokkinos2019iterative,ehret2019joint,liu2020joint,guo2021joint}. 

Gharbi \emph{et al.}~\cite{gharbi2016deep} showed that using more challenging patches for training can reduce the moir\'{e} artifacts in JDD. Then, more complex methods have been to obtain better performance, \eg, two-stage network~\cite{qian2019trinity}, auto-encoder architecture~\cite{henz2018deep} and iterative structure~\cite{kokkinos2019iterative}.
The mosaic-to-mosaic framework~\cite{ehret2019joint} was proposed to improve the demosaicking performance on real-world images by finetuning the network with burst images. Since in the Bayer pattern of camera raw images, the green channel has twice the sampling rate of red/blue channels and has higher signal-to-noise ratio, the green channels were utilized to guide the upsampling process~\cite{liu2020joint}, the feature extraction~\cite{guo2021joint} and offset estimation between frames~\cite{guo2021joint}. Previous methods mostly perform JDD on single CFA raw images and achieve limited performance on real-world CFA image with high noise level. Even a method has been proposed to use burst raw image for the JDD task \cite{guo2021joint}, its alignment module has small receptive field and cannot obtain visual-pleasing results on images with large shift. In this work, we propose a new two-stage framework, which obtains great improvement over \cite{guo2021joint} with similar running time.

\subsection{Multi-frame and Burst Image Restoration}
Multi-frame image restoration has been widely studied in literature~\cite{ehret2019model,xue2019video,mildenhall2018burst,xu2019learning,marinvc2019multi,xia2019basis,liu2017robust,tian2018tdan,wang2019edvr,chan2020basicvsr,yue2020supervised,isobe2020video}, aiming to reproduce a photo with better quality than that estimated from a single image. The major challenge of multi-frame image restoration lies in how to compensate for the motion between frames. Some popular solutions employ optical flow~\cite{ehret2019model,xue2019video} and spatially varying kernel estimation~\cite{mildenhall2018burst,xu2019learning,marinvc2019multi,xia2019basis}. However, the performance of these methods is highly affected by large motion and severe noise.


Recently, performing implicit frame alignment in feature domain has achieved state-of-the-art performance on video super-resolution~\cite{liu2017robust,tian2018tdan,wang2019edvr,chan2020basicvsr}, video denoising~\cite{yue2020supervised} and JDD-B~\cite{guo2021joint}. In such methods, the offset~\cite{tian2018tdan,wang2019edvr} or optical flow~\cite{chan2020basicvsr} is estimated from the deep features extracted from neighboring frames, and the deformable convolution~\cite{dai2017deformable} or wrap operator is used to compensate for shifts in feature domain. In \cite{wang2019edvr,yue2020supervised,guo2021joint}, pyramidal processing is employed in the offset estimation to deal with complex motions. However, we found that, due to the limited receptive field, implicit alignment in feature domain can only achieve limited performance for sequences with large shift. To solve this problem, we design a differentiable two-stage alignment scheme and demonstrate its effectiveness for JDD-B.

\begin{table*}[!tbp]
\setlength{\abovecaptionskip}{0.0cm}
\setlength{\belowcaptionskip}{0.0cm}
\footnotesize
\centering
\caption{Quantitative comparison of different JDD-B approaches on the REDS4 dataset. Following the experiment setting of \cite{mildenhall2018burst,xu2019learning}, ``Low" and ``High" noise levels are corresponding to $\sigma_s = 2.5\times 10^{-3}$, $\sigma_r = 10^{-2}$ and $\sigma_s = 6.4\times 10^{-3}$, $\sigma_r = 2\times 10^{-2}$, respectively.} \vspace{0.2cm}
\label{synthREDS4}
\begin{tabular}{c c c c c c c c c}
\toprule
Noise Level & Clip Name &KPN+DMN &EDVR+DMN &RviDeNet+DMN &EDVR* &RviDeNet* &GCP-Net &Ours \\
\midrule
\multirow{5}{*}{Low} 
& \emph{Clip000} &29.85/0.8398 &32.15/0.8984 &32.38/0.9075 &32.44/0.9072 &33.58/0.9285 &34.96/0.9412 &\textbf{35.07/0.9501}\\
& \emph{Clip011} &32.01/0.8496 &34.29/0.9030 &34.39/0.9019 &34.32/0.9017 &34.63/0.9044 &36.21/0.9288 &\textbf{36.79/0.9356} \\
& \emph{Clip015} &33.46/0.8770 &35.42/0.9074 &35.75/0.9169 &35.54/0.9160 &36.98/0.9327 &37.74/0.9403 &\textbf{38.13/0.9456}\\
& \emph{Clip020} &31.29/0.8613 &33.56/0.9071 &33.81/0.9165 &33.86/0.9179 &34.23/0.9230 &35.92/0.9428 &\textbf{36.38/0.9479}\\
\cmidrule{2-9}
& Average  &31.65/0.8570 &33.85/0.9039 &34.08/0.9107 &34.02/0.9105 &34.86/0.9221 &36.20/0.9383 &\textbf{36.59/0.9448} \\

\midrule
\multirow{5}{*}{High} 
& \emph{Clip000} &27.47/0.7437 &30.18/0.8517 &30.37/0.8532 &30.31/0.8539 &31.29/0.8803 &32.57/0.9147 &\textbf{32.75/0.9174}  \\
& \emph{Clip011} &29.64/0.7886 &32.26/0.8611 &32.50/0.8639 &32.51/0.8643 &32.50/0.8710 &34.20/0.8972 &\textbf{34.74/0.9059} \\
& \emph{Clip015} &31.21/0.8310 &34.01/0.8893 &34.10/0.8909 &34.01/0.8919 &34.90/0.9068 &35.94/0.9225 &\textbf{36.06/0.9247} \\
& \emph{Clip020} &28.66/0.7938 &31.65/0.8780 &31.74/0.8807 &31.44/0.8757 &31.82/0.8822 &33.61/0.9154 &\textbf{34.17/0.9233} \\
\cmidrule{2-9}
& Average  &29.24/0.7893 &32.02/0.8700 &32.18/0.8722 &32.06/0.8715 &32.62/0.8850 &34.08/0.9124 &\textbf{34.43/0.9178}\\

\toprule
\end{tabular} 
\vspace{-0.2cm}
\end{table*}

\section{Methodology}
\label{sec:methods}
\subsection{Motivation and Network Structure}
The purpose of JDD-B task is to reconstruct clean RGB image $x$ from a burst of noisy CFA images $\mathcal{Y} = \{ y_t \}_{t=1}^{N}$ and their corresponding noise maps $\mathcal{M} = \{ m_t \}_{t=1}^{N}$. 
However, previous SOTA multi-frame methods~\cite{wang2019edvr,yue2020supervised,guo2021joint} have limited ability to address large pixel displacement by the small receptive fields. Thus, we propose a differentiable two-stage alignment framework to increase the receptive field of alignment without increasing the amount of calculation. Our framework is illustrated in Fig.~\ref{figOverall}, which divides the difficult large shift compensation problem into two relative easier sub-problems, \ie, coarse alignment (CA) and refined alignment (RA). 

The CA module needs to compensate large shift roughly using small computational cost. We choose block matching (BM) based method and propose a differentiable progressive block matching method to estimate the offset on low-resolution (LR) features and output the coarsely aligned burst images. The RA module is then developed, which uses deformable convolution (DConv) to pixel-wise align the burst images in high-resolution (HR) feature domain. Finally, the fusion module estimates the clean full color image of the reference frame (denoted as frame $r$) by using the aligned image features. The details of each module are described in the following sections.

\subsection{Coarse Alignment in Patch Level}
The coarse alignment (CA) module aims to efficiently estimate the large offsets between frames. To meet the requirements of handling burst images with complex motion at 4K resolution, the CA module needs to have a large receptive field using small cost.
Also in model training, to suppress the effect of small noise, the GT images are usually obtained by down-sampling high-resolution images, which also reduces the range of motion. Thus we choose to use BM to perform coarse alignment, which can easily mitigate this gap between training and testing stages by increasing the search region.
To overcome the non-differentiability caused by BM and reduce computational cost, we propose a differentiable progressive block matching (DPBM) method. 

\noindent\textbf{Differentiable BM.} We first introduce the differentiable BM process. For a patch $P_r$ in the reference frame $r$ and a query patch $(P_{t,i})_{i\in I}$ in another frame $t$ with indices $I=\{1,...,M\}$, the purpose of BM is to find the best matched patch in $(P_{t,i})_{i\in I}$ with the target patch $P_r$. The normalized mean-absolute (NMA) distance is used as the matching criterion. The NMA distance between image patches $P_{t,i}$ and $P_r$ can be written as:
\begin{equation}
    d(P_{t,i},P_r) = \frac{\mathbb{E}[\vert P_{t,i} - P_r \vert]}{\sqrt{\sum_{i=1}^M \mathbb{E}[\vert P_{t,i} - P_r \vert]^2}}.
\end{equation}
For traditional BM, the patch with minimum distance is chosen as the best matched patch, which is denoted as $P_{t,BM}$. The minimum distance is obtained by sorting $d(P_{t,i},P_r)_{i\in I}$ in ascending order, which is not differentiable. To overcome this problem, we refer to~\cite{goldberger2004neighbourhood,jang2016categorical,plotz2018neural} and utilize continuous deterministic relaxation of BM. The weighting factor $w$ of $d(P_{t,i},P_r)_{i\in I}$ can be calculated as:
\begin{equation}
    w_i = \frac{\exp{(-d(P_{t,i},P_r)/T)}}{\sum_{i'\in I} \exp{(-d(P_{t,i'},P_r)/T)}},
\label{gumbel}
\end{equation}
where $T$ is the temperature and $w_i$ becomes one-hot vector when $T$  approaches 0. We now can obtain $P_{t,BM}$ by using $P_{t,BM} = \sum_{i\in I} w_iP_{t,i}$.
As discussed in \cite{jang2016categorical}, using small $T$ can make $w$ closer to one-hot, but it will result in a larger gradient and cause the training process to be unstable. While using large $T$, $w$ becomes smooth but easy to train. Thus we regard training differentiable BM as a soft to hard process, where $T$ is initialized as $1\times 10^{-2}$ and then reduced to $1\times 10^{-3}$ in our experiments. 

Different from  the differnetiable BM in~\cite{plotz2018neural}, here we calculate the distance between patches on LR features obtained by a lightweight network, which is updated by the gradient only from differentiable BM. We further design some constrains for DPBM to stablize the training process, which will be introduced in Sec.~\ref{DPBMLoss}. 

\noindent\textbf{DPBM.}
The detailed process of DPBM is illustrated in Fig.~\ref{figPBM}. Firstly, to reduce the computational cost and relieve the impact of noise interference, we perform differentiable BM on LR features which are obtained by a lightweight downsampling network. The network contains three $3\times 3$ convolution (conv) layers with 16 channels and two transpose conv layers to downsample images into learned features with $1/4$ scale. For the LR feature patch $P_{r,p}$ centered at position $p$ in the reference frame $r$, the search region in its closest target frame, denoted as frame $r+1$, is set to $\{p\pm \Delta p_{cmax}\}$, where $\Delta p_{cmax}$ is the max search range. We firstly find an approximately matched position in $\{p\pm \Delta p_{cmax}\}$ with a stride $s>1$ using differentiable BM and denote the matched position as $p_a$. Then we find a more accurate matching patch in $\{p_a\pm s\}$ with stride 1 and denote the matched position as $p_m$. The shift between frames $r$ and $r+1$ for patch $P$ is obtained as $\Delta p_{r+1} = p_m - p$.

For target frame $r+2$, the search region is set to $\{p + \Delta p_{r+1} \pm \Delta p_{cmax}\}$, which is updated using the offset estimated from frame $r+1$. Usually, motion between frames has continuity, and updating the search region using temporal information can more effectively locate the matched patch in long-term frames. In the cases where the motion between frames changes abruptly, our DPBM is equivalent to the standard BM. Note that the DPBM is performed on 1/4 scale LR features and the estimated offsets should be rescaled to the original image resolution when aligning the full resolution images. 

\begin{figure*}[!t]
\setlength{\abovecaptionskip}{0.0cm}
\setlength{\belowcaptionskip}{-0.cm}
\centering
\subfloat{
\begin{minipage}[t]{0.24\textwidth}
\centering
\includegraphics[width=1\textwidth]{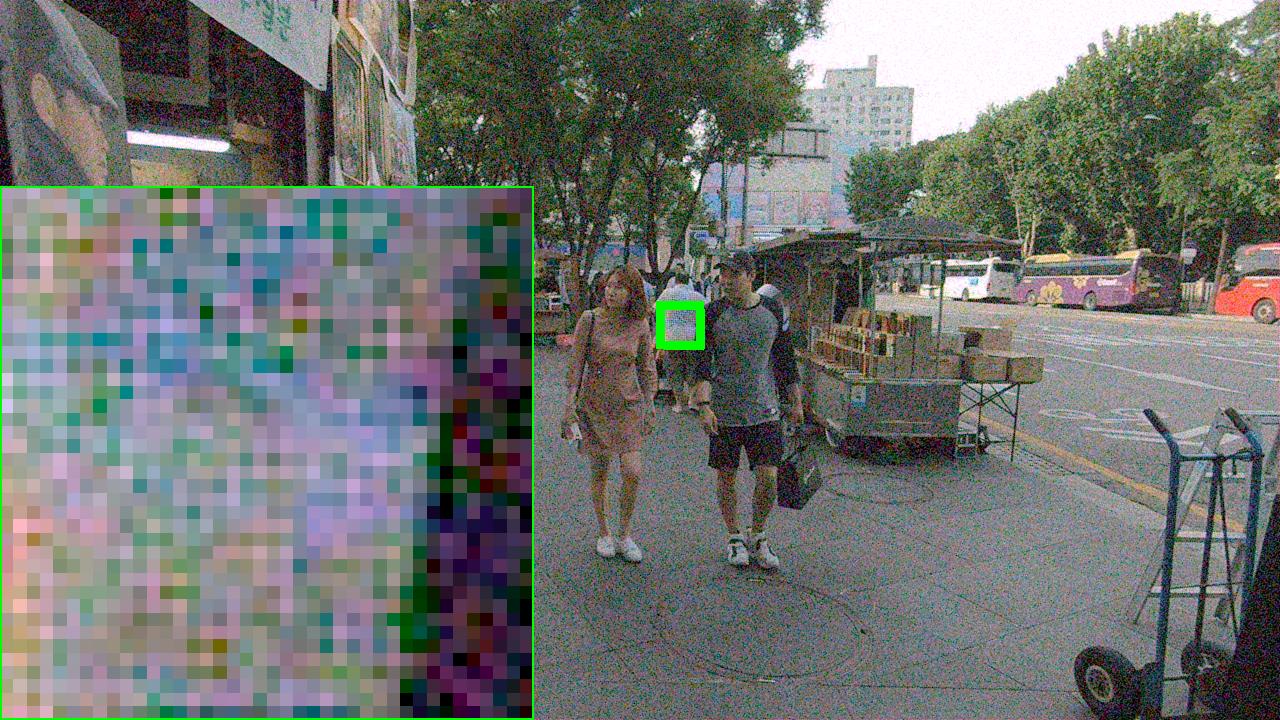}
{\footnotesize  (a) Noisy image}
\end{minipage}\hspace{0.05pt}
\begin{minipage}[t]{0.24\textwidth}
\centering
\includegraphics[width=1\textwidth]{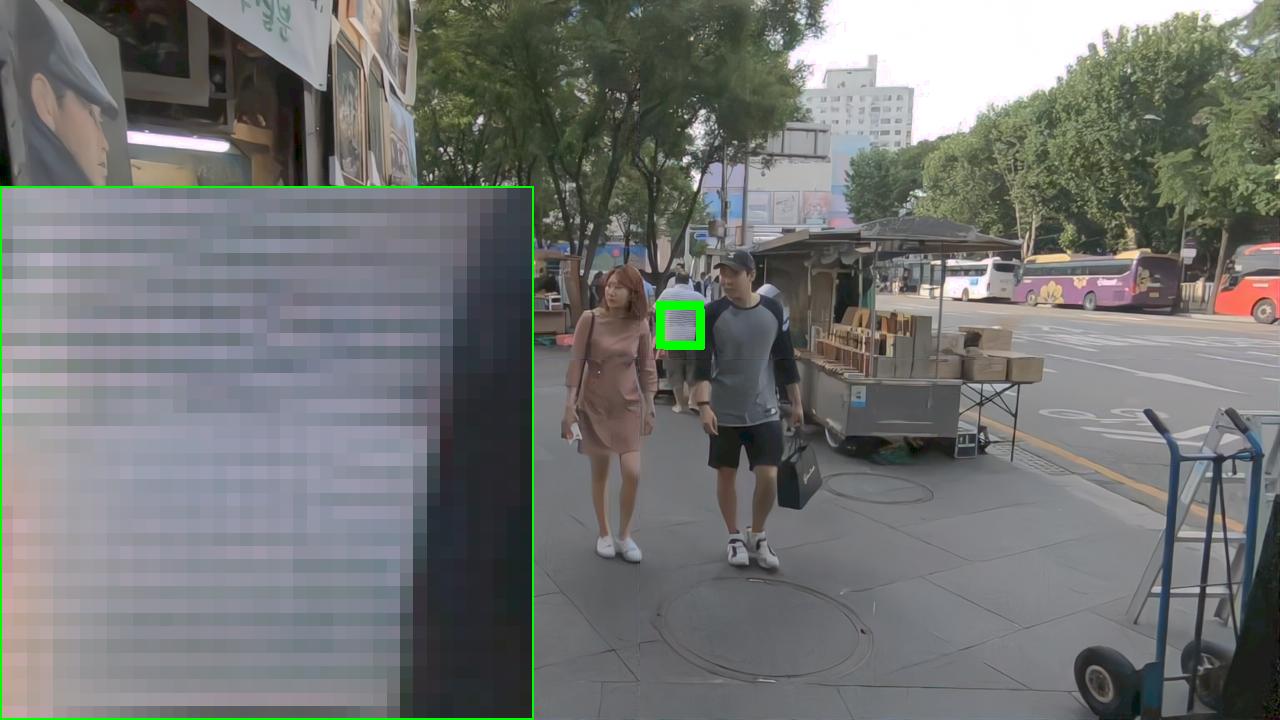}
{\footnotesize  (b) EDVR+DMN}
\end{minipage}\hspace{0.05pt}
\begin{minipage}[t]{0.24\textwidth}
\centering
\includegraphics[width=1\textwidth]{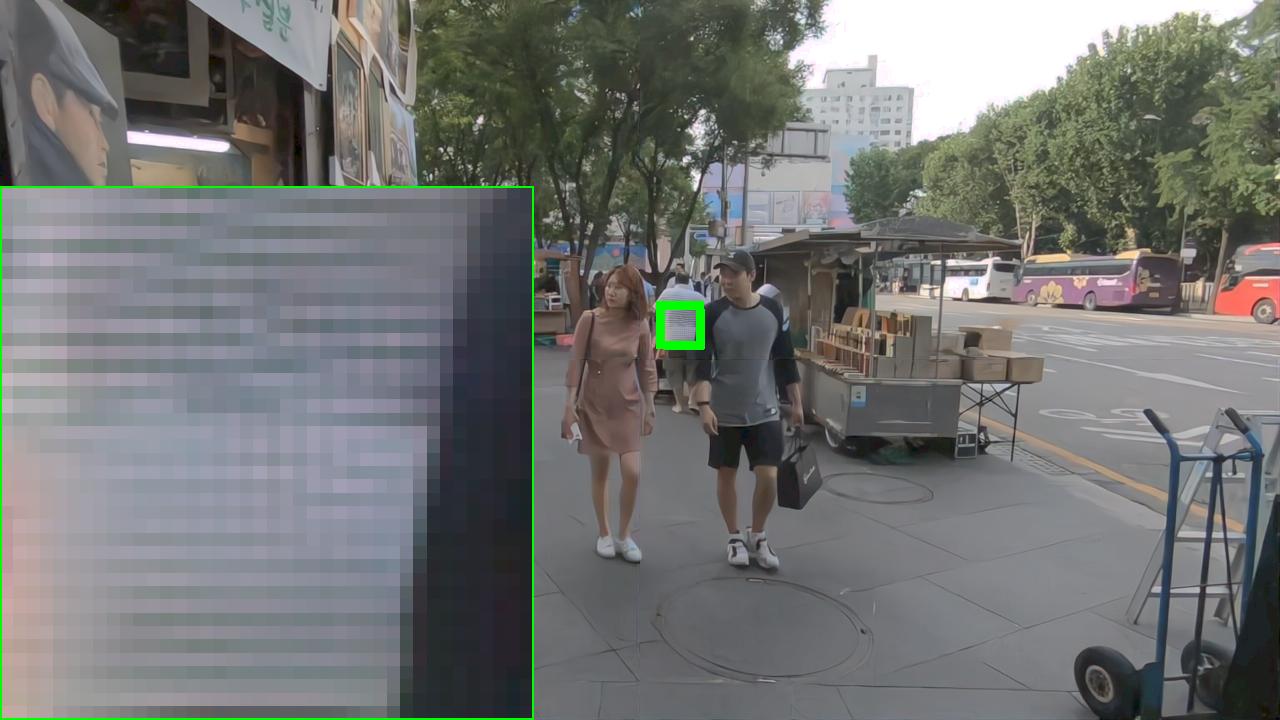}
{\footnotesize  (c) RviDeNet+DMN}
\end{minipage}\hspace{0.05pt}
\begin{minipage}[t]{0.24\textwidth}
\centering
\includegraphics[width=1\textwidth]{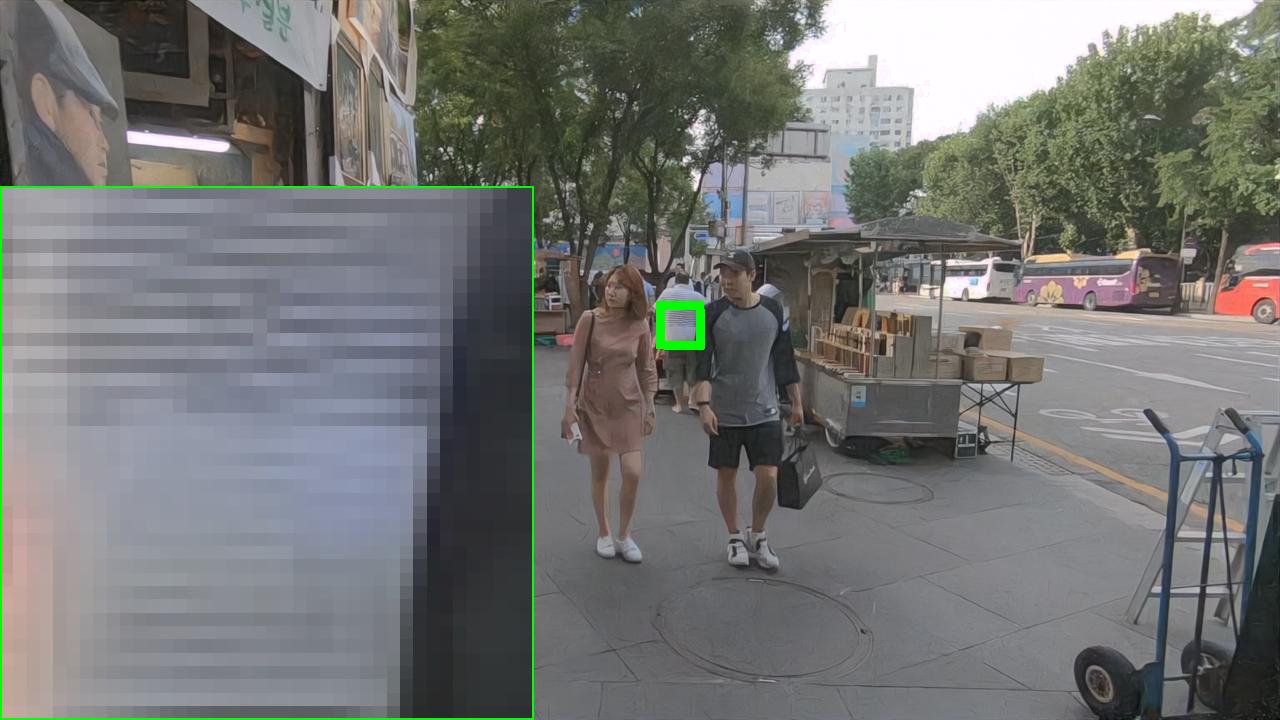}
{\footnotesize  (d) EDVR*}
\end{minipage}\hspace{0.05pt}
}

\subfloat{
\begin{minipage}[t]{0.24\textwidth}
\centering
\includegraphics[width=1\textwidth]{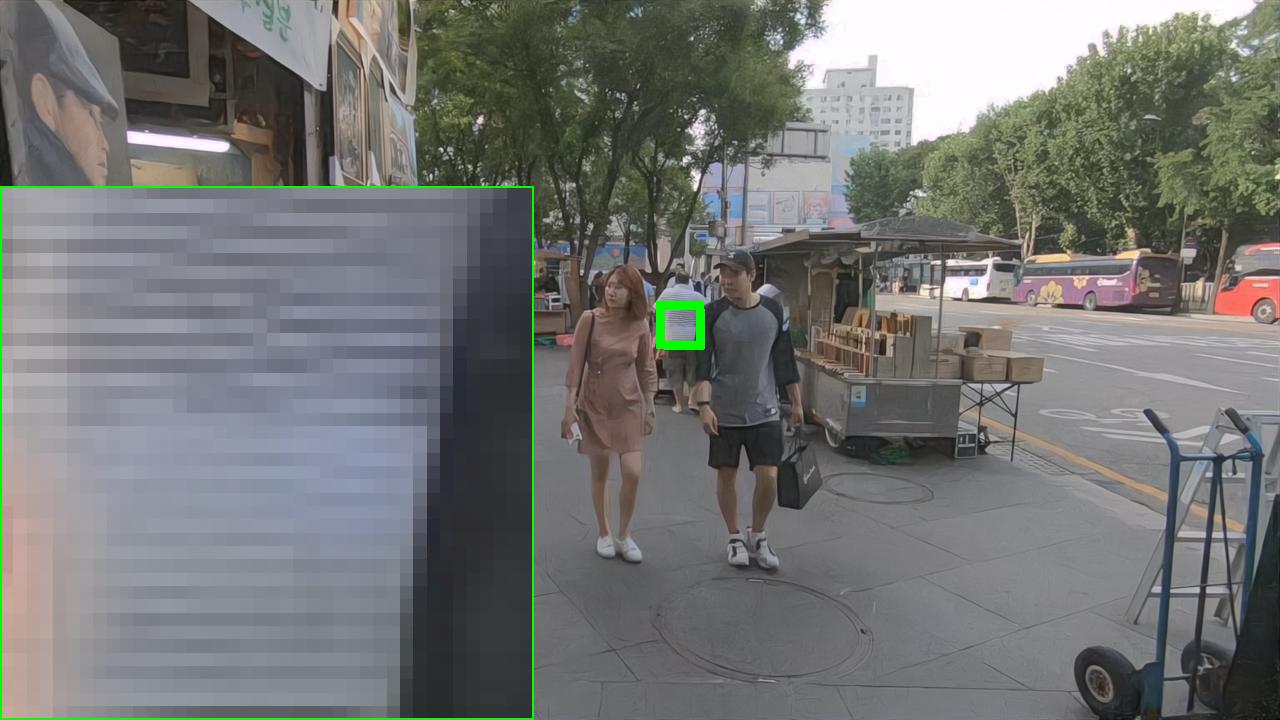}
{\footnotesize  (e) RviDeNet*}
\end{minipage}\hspace{0.05pt}
\begin{minipage}[t]{0.24\textwidth}
\centering
\includegraphics[width=1\textwidth]{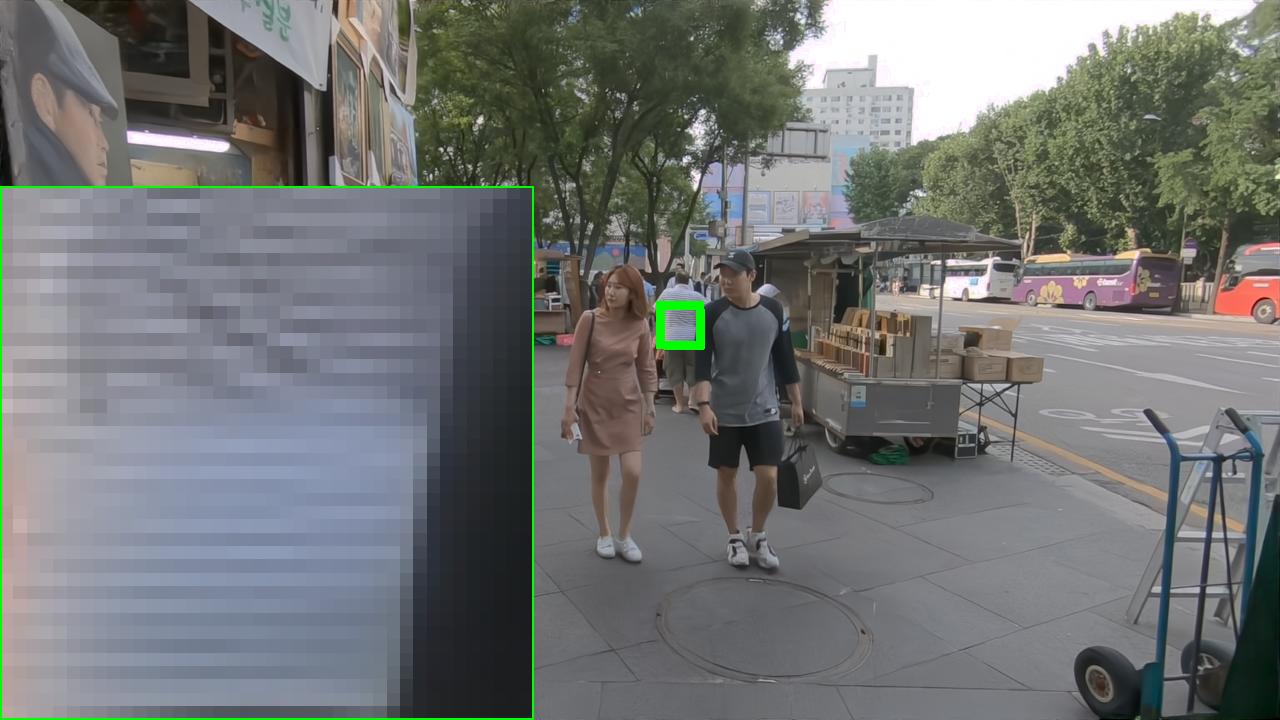}
{\footnotesize  (f) GCP-Net}
\end{minipage}\hspace{0.05pt}
\begin{minipage}[t]{0.24\textwidth}
\centering
\includegraphics[width=1\textwidth]{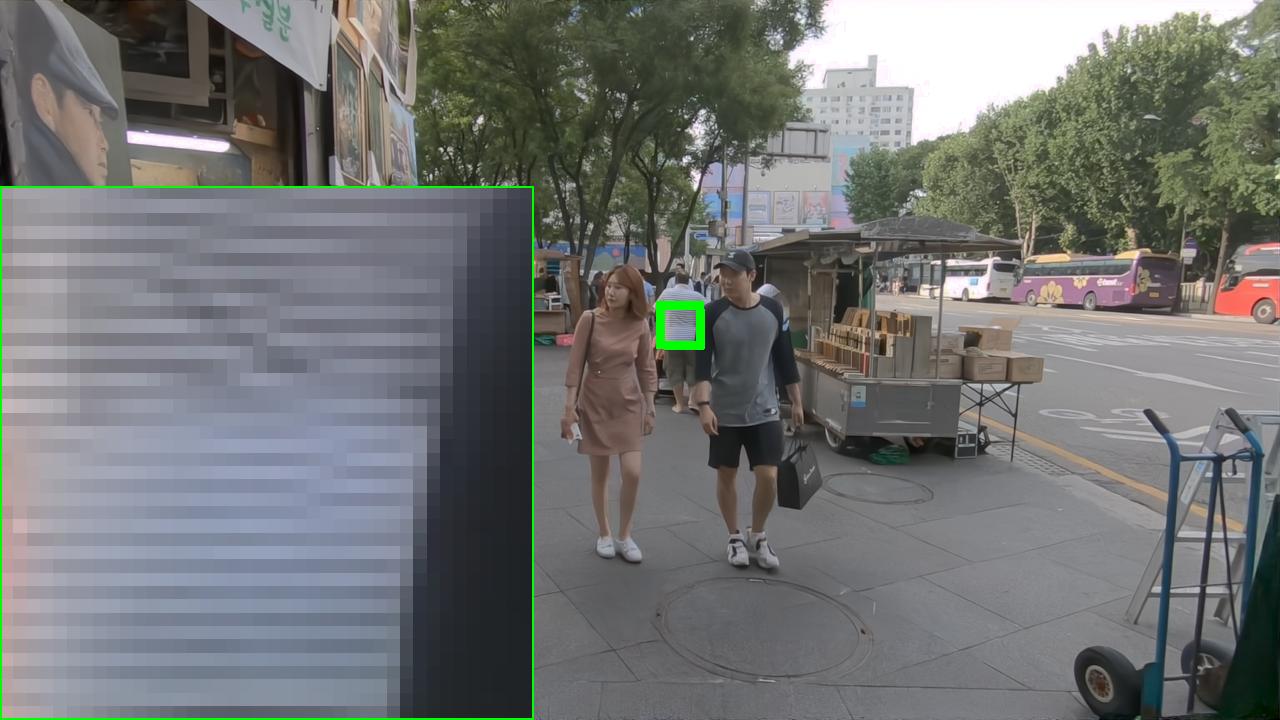}
{\footnotesize  (g) Ours}
\end{minipage}\hspace{0.05pt}
\begin{minipage}[t]{0.24\textwidth}
\centering
\includegraphics[width=1\textwidth]{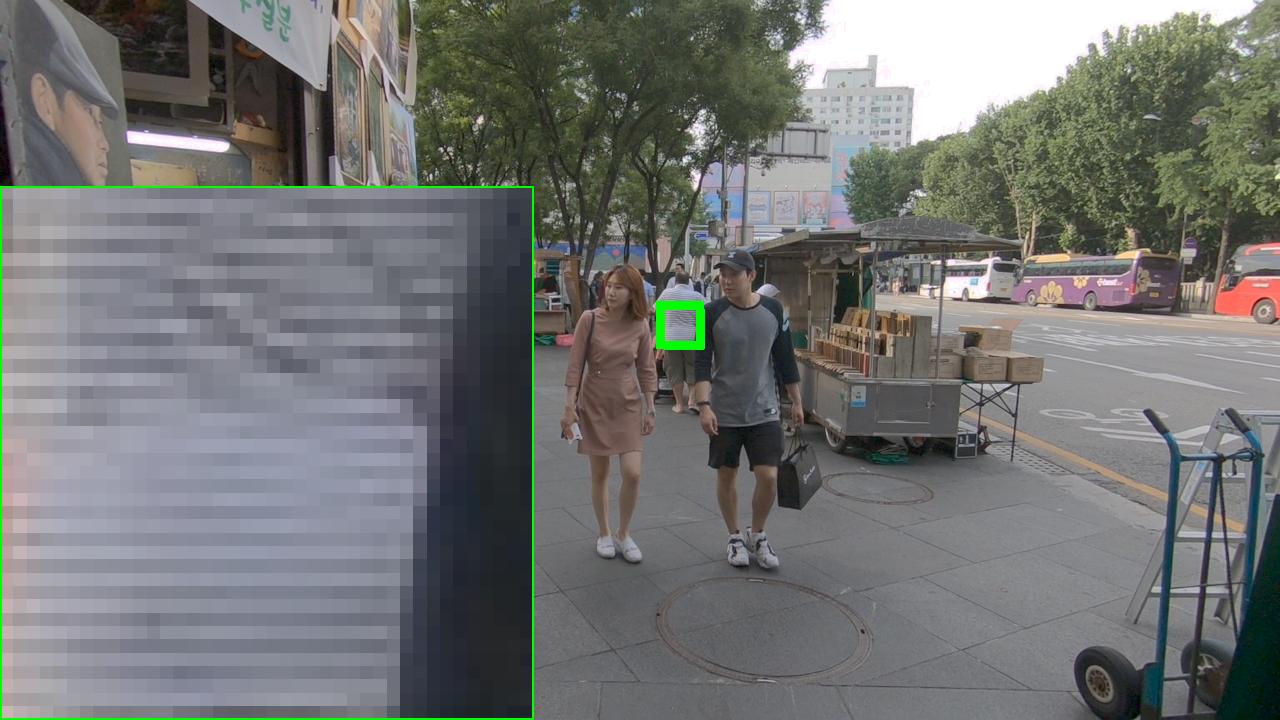}
{\footnotesize  (h) GT}
\end{minipage}\hspace{0.05pt}
}

\caption{JDD-B results on \emph{Clip 020} of the REDS4 dataset by different methods.}
\label{figRED}
\vspace{-0.2cm}
\end{figure*}

\subsection{Refined Alignment in Pixel Level}
The refined alignment module aims to perform accurate pixel-wise alignment on the coarsely aligned frames. The deformable alignment with DConv~\cite{dai2017deformable} has proved its success in feature alignment for various video processing tasks~\cite{tian2018tdan,wang2019edvr,yue2020supervised,guo2021joint}. Thus we choose to use implicit feature alignment in our refined alignment module.

The deep features are extracted from the burst images at the original resolution. The alignment of features of the reference frame and a target frame, denoted by $F_r$ and $F_{r+1}$, is described as follows. Firstly, the offset is estimated from the deep features by using $ \Delta p_{r+1} = f([F_r,F_{r+1}])$,
where $[\cdot,\cdot]$ is the concatenation operator and $f$ refers to the combination of several Conv layers and nonlinear functions (\emph{e.g.}, LReLU). Then the aligned feature $F_{r+1}'$ can be obtained by performing DConv with the estimated offset:
\begin{equation}
    F_{r+1}' = \text{DConv}(F_{r+1}, \Delta p_{r+1}).
\end{equation}
Similar to \cite{wang2019edvr}, we also adopt the pyramid DConv alignment strategy to improve the alignment performance.

\subsection{Aligned Feature Fusion}
In burst image restoration, the merging of aligned features is another important step. There are three typical solutions, \ie, non-directional fusion~\cite{wang2019edvr,li2020mucan,isobe2020video,yue2020supervised},  unidirectional fusion~\cite{sajjadi2018frame,isobe2020video} and bidirectional fusion~\cite{huang2015bidirectional,huang2017video,chan2020basicvsr}. 
Compared with non-directional and unidirectional fusion, bidirectional fusion can utilize long-term memory and information from both forward and backward directions~\cite{chan2020basicvsr}. Thus, we design a bi-directional gate recurrent unit (Bi-GRU) as the core component of our fusion module. 

The input of our fusion module is the aligned feature frames $\{ F_t'\}_{t=1}^N$.  For the reference frame $r$, the features are propagated with forward ($h_r^f$) and backward ($h_r^b$) information, which can be calculated as:
\begin{equation}
    h_r^f = f_{gru}(F_r, h_{r-1}^f), \quad h_r^b = f_{gru}(F_r, h_{r+1}^b),
\label{equ:bigruagg}
\end{equation}
where $f_{gru}$ refers to the gate recurrent unit (GRU). The outputs of forward pass and backward pass are then concatenated as the input for clean full-color image reconstruction. A typical 3-scale UNet with two skip connections is used as the image reconstruction network. More details of network structure are provided in the supplementary file.

\begin{table*}[!tbp]
\setlength{\abovecaptionskip}{0.0cm}
\setlength{\belowcaptionskip}{0.0cm}
\footnotesize
\centering
\caption{Quantitative comparison on the Videezy4K dataset. The noise level is set as $\sigma_s = 6.4\times 10^{-3}$, $\sigma_r = 2\times 10^{-2}$.} \vspace{0.1cm}
\label{synthVideezy}
\begin{tabular}{c c c c c c c c c c c c c c}
\toprule
Methods &\emph{00} &\emph{01} &\emph{02} &\emph{03} &\emph{04} &\emph{05} &\emph{06} &\emph{07} &\emph{08} &\emph{09} &\emph{10} &\emph{11} &Average \\
\midrule
\multirow{2}{*}{EDVR*}  &36.07 &38.45 &40.07 &40.84 &37.82 &33.93 &36.96 &35.71 &37.02 &34.86 &37.60 &36.99 &37.19\\
&0.9406 &0.9583 &0.9652 &0.9721 &0.9511 &0.9207 &0.9454 &0.9310 &0.9419 &0.9299 &0.9387 &0.9367 &0.9443 \\
\midrule
\multirow{2}{*}{RviDeNet*} &36.22 &38.56 &40.78 &40.42 &38.34 &34.30 &37.76 &36.40 &37.23 &34.83 &38.04 &37.16 &37.50\\
&0.9490 &0.9657 &0.9738 &0.9789 &0.9582 &0.9271 &0.9543 &0.9399 &0.9448 &0.9385 &0.9432 &0.9396 &0.9510\\
\midrule
\multirow{2}{*}{GCP-Net} &35.51 &39.17 &41.32 &40.85 &38.74 &34.83 &38.50 &37.12 &37.69 &35.67 &38.21 &37.78 &37.94 \\
&0.9306 &0.9683 &0.9755 &0.9646 &0.9653 &0.9365 &0.9650 &0.9530 &0.9482 &0.9458 &0.9473 &0.9424 &0.9535 \\
\midrule
\multirow{2}{*}{Ours} &\textbf{37.75} &\textbf{39.61} &\textbf{41.71} &\textbf{42.42} &\textbf{39.30} &\textbf{36.09} &\textbf{39.24} &\textbf{37.21} &\textbf{38.76} &\textbf{35.94} &\textbf{38.64} &\textbf{38.27} &\textbf{38.74} \\
&\textbf{0.9578} &\textbf{0.9737} &\textbf{0.9809} &\textbf{0.9855} &\textbf{0.9698} &\textbf{0.9532} &\textbf{0.9741} &\textbf{0.9594} &\textbf{0.9604} &\textbf{0.9491} &\textbf{0.9515} &\textbf{0.9514} &\textbf{0.9639} \\

\toprule
\end{tabular} 
\vspace{-0.2cm}
\end{table*}
\begin{figure*}[!h]
\setlength{\abovecaptionskip}{0.0cm}
\setlength{\belowcaptionskip}{-0.cm}
\centering

\begin{minipage}[b]{1.0\textwidth}
\centering
    \begin{minipage}[b]{0.16\textwidth}
    \centering
    \includegraphics[width=1\textwidth]{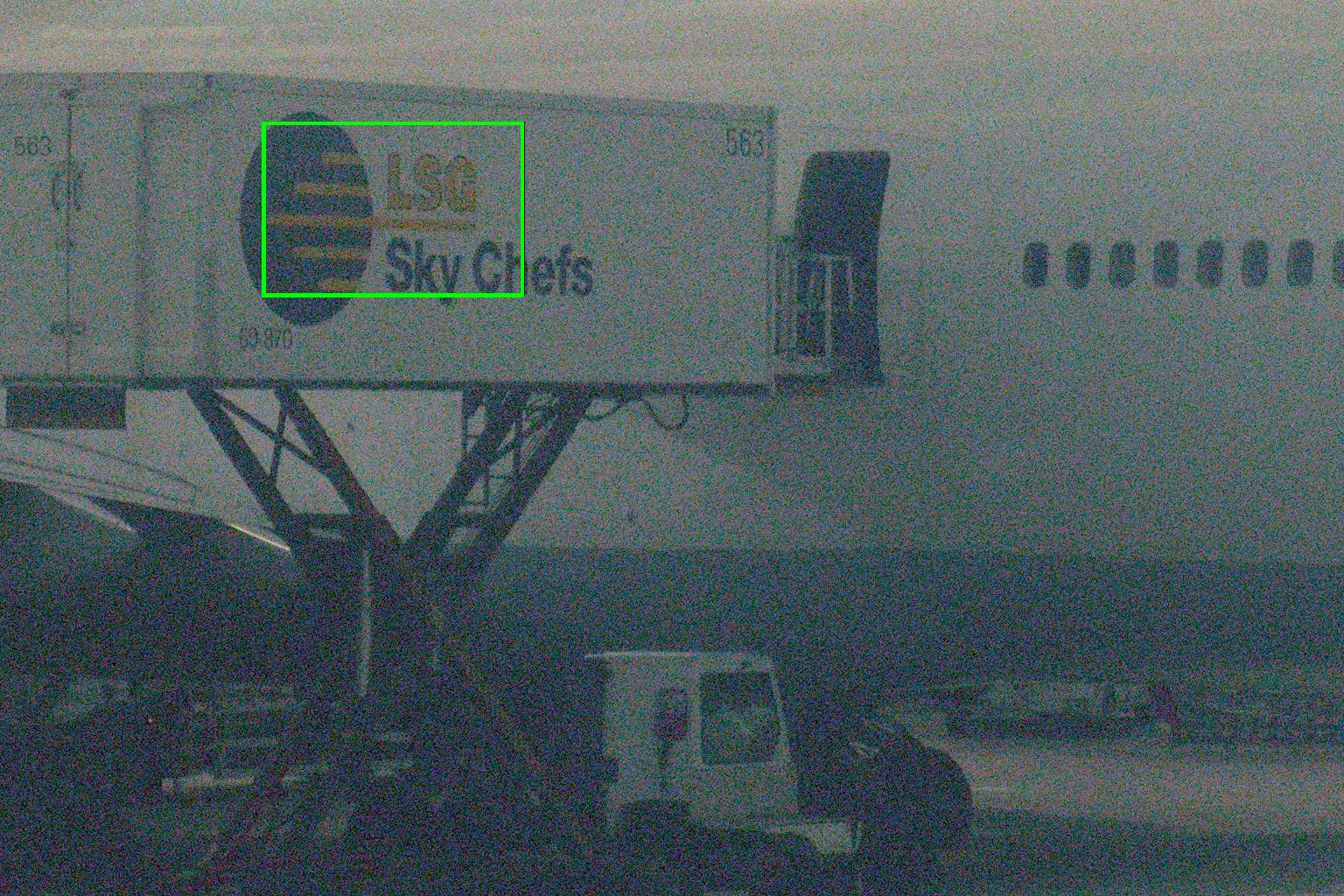}
    \end{minipage} 
    \begin{minipage}[b]{0.16\textwidth}
    \centering
    \includegraphics[width=1\textwidth]{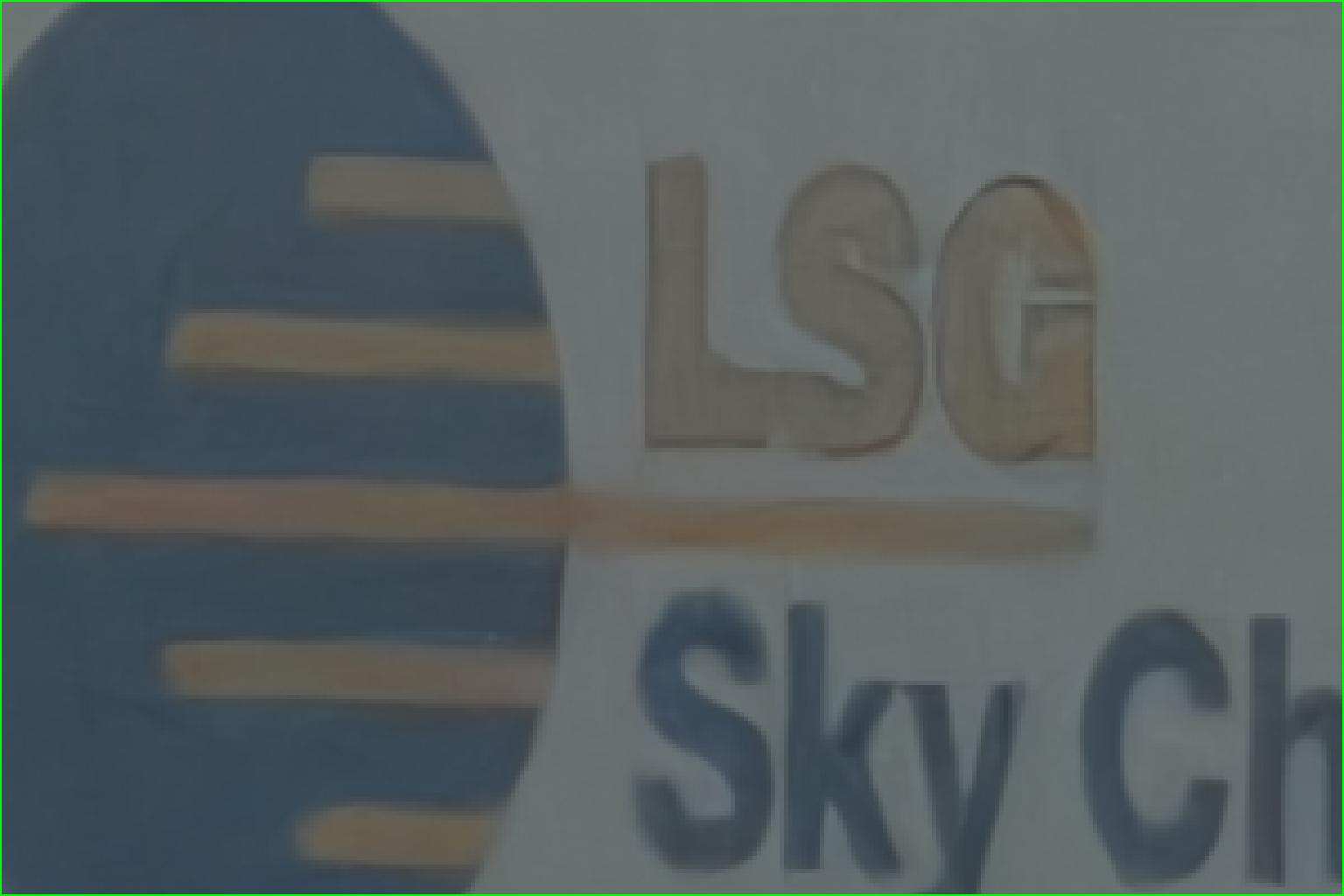}
    \end{minipage} 
    \begin{minipage}[b]{0.16\textwidth}
    \centering
    \includegraphics[width=1\textwidth]{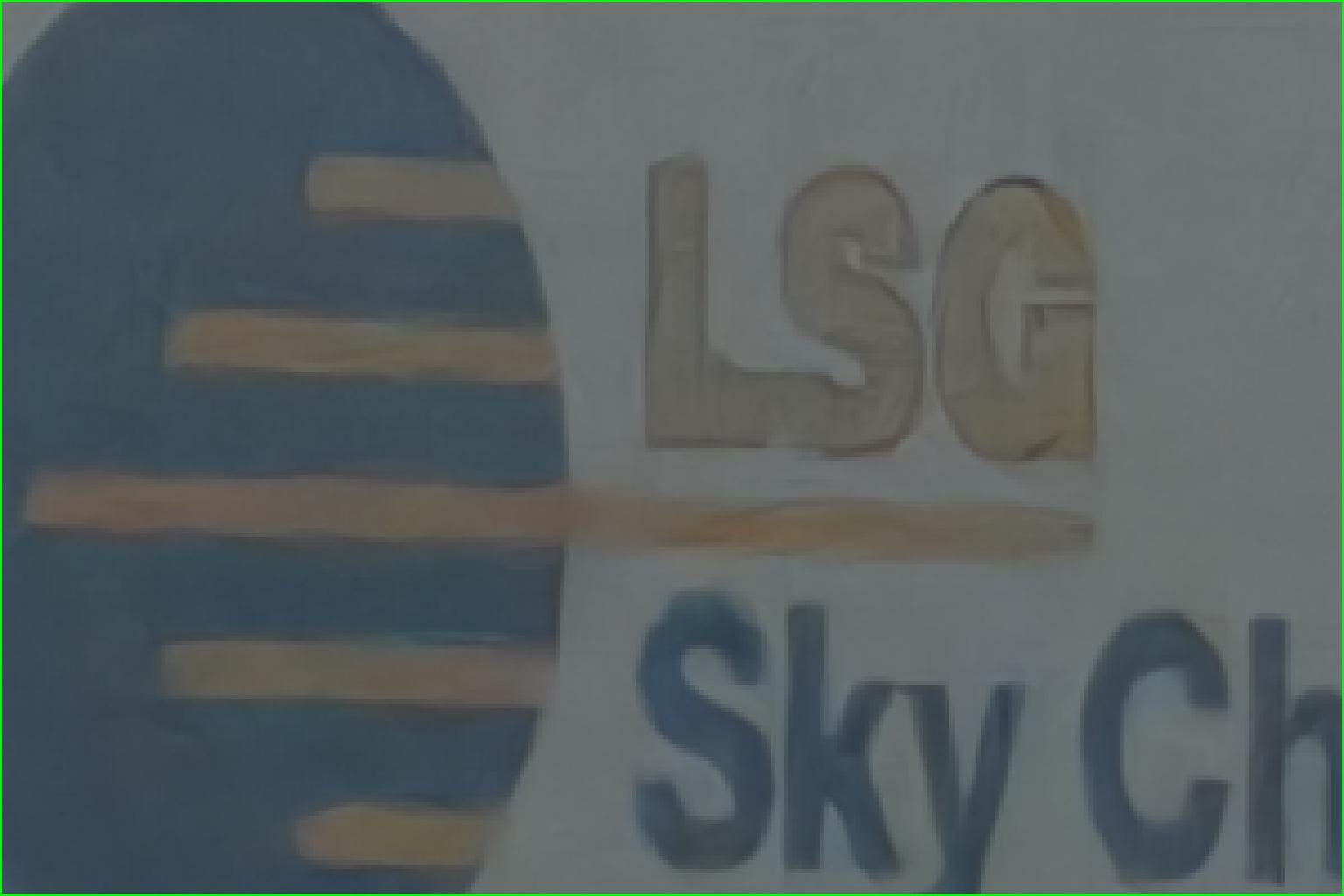}
    \end{minipage} 
    \begin{minipage}[b]{0.16\textwidth}
    \centering
    \includegraphics[width=1\textwidth]{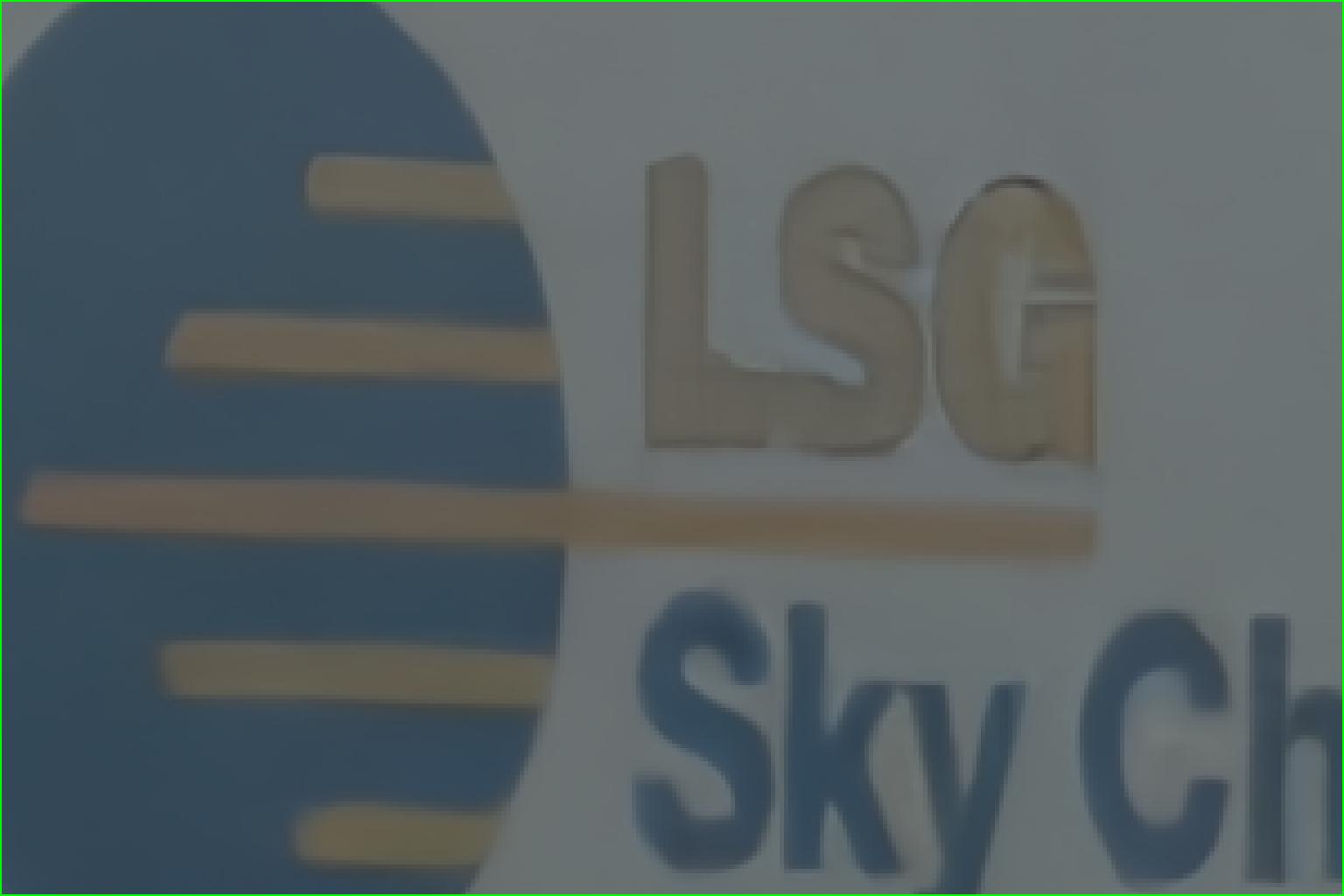}
    \end{minipage} 
    \begin{minipage}[b]{0.16\textwidth}
    \centering
    \includegraphics[width=1\textwidth]{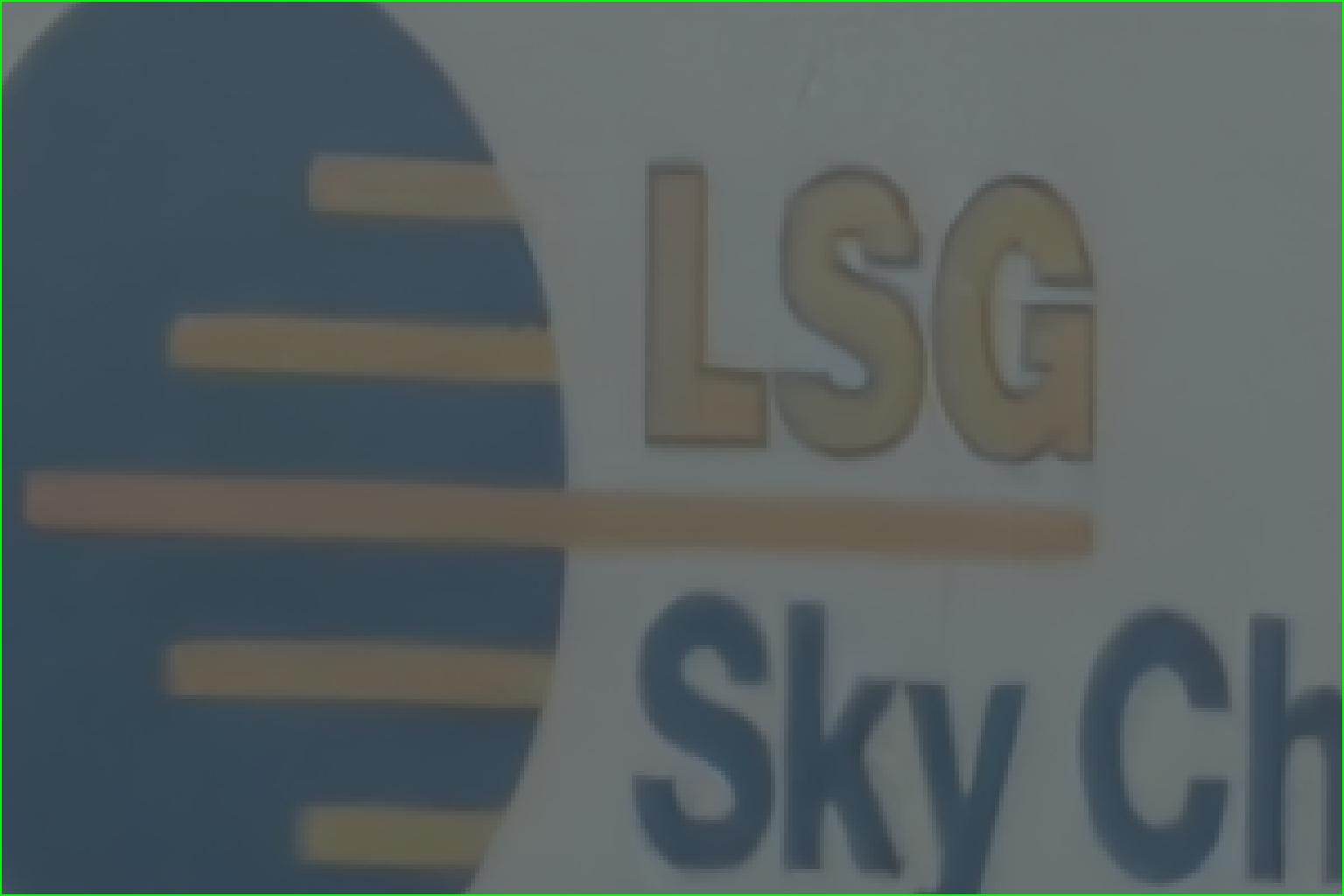}
    \end{minipage} 
    \begin{minipage}[b]{0.16\textwidth}
    \centering
    \includegraphics[width=1\textwidth]{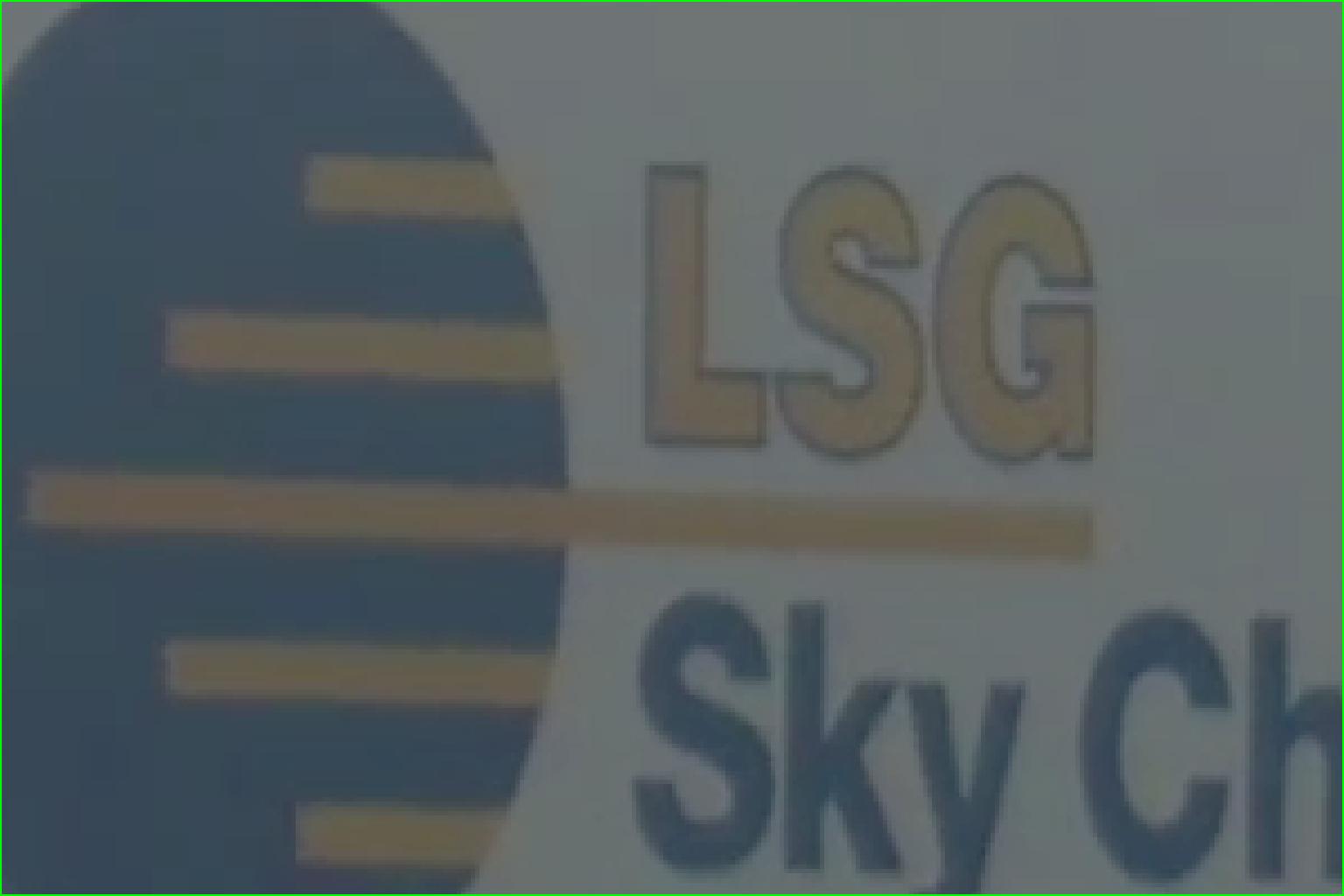}
    \end{minipage} 
\end{minipage}
\vspace{0.15cm}

\begin{minipage}[b]{1.0\textwidth}
\centering
    \begin{minipage}[b]{0.16\textwidth}
    \centering
    \includegraphics[width=1\textwidth]{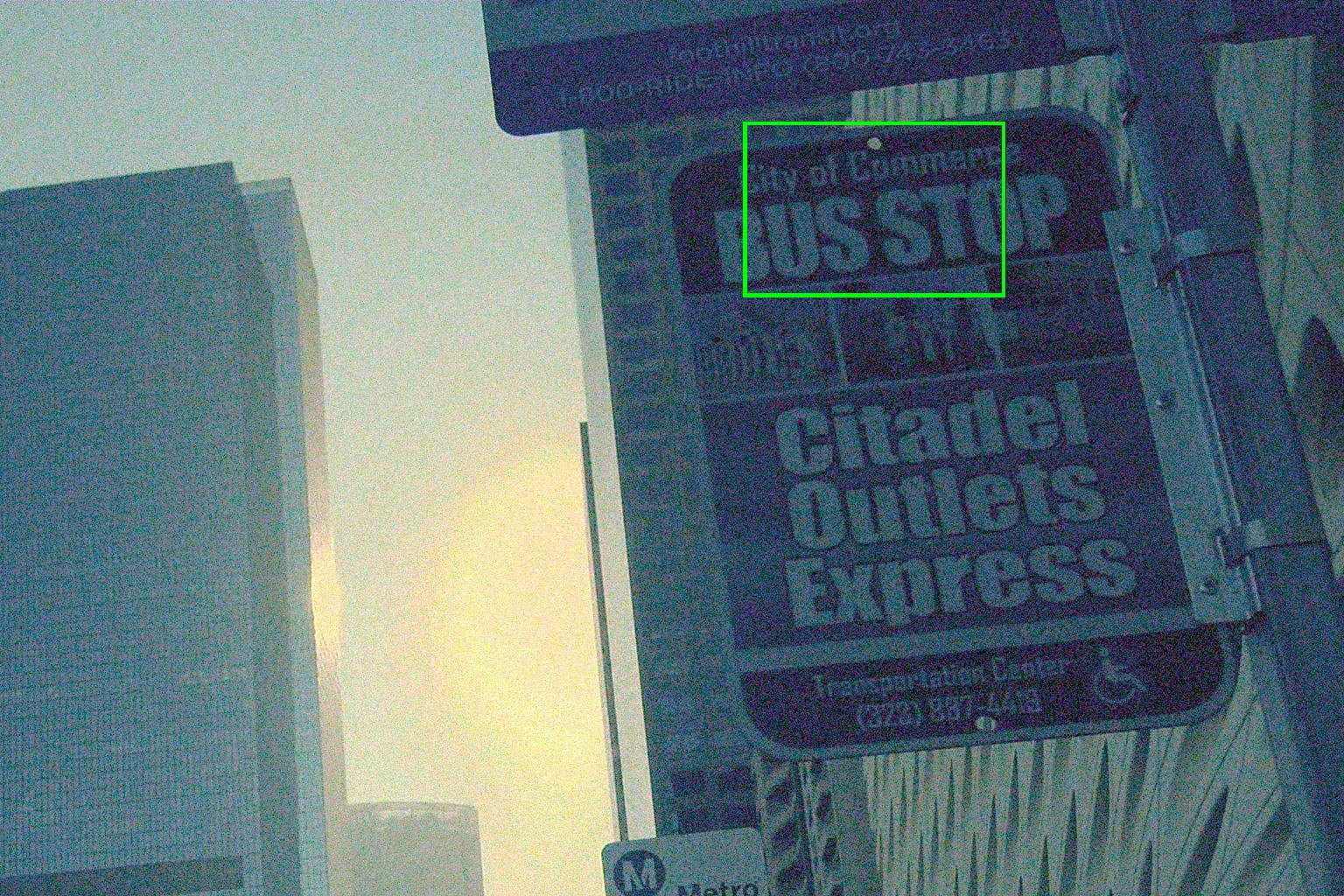}
    {\small (a) Noisy image}
    \end{minipage} 
    \begin{minipage}[b]{0.16\textwidth}
    \centering
    \includegraphics[width=1\textwidth]{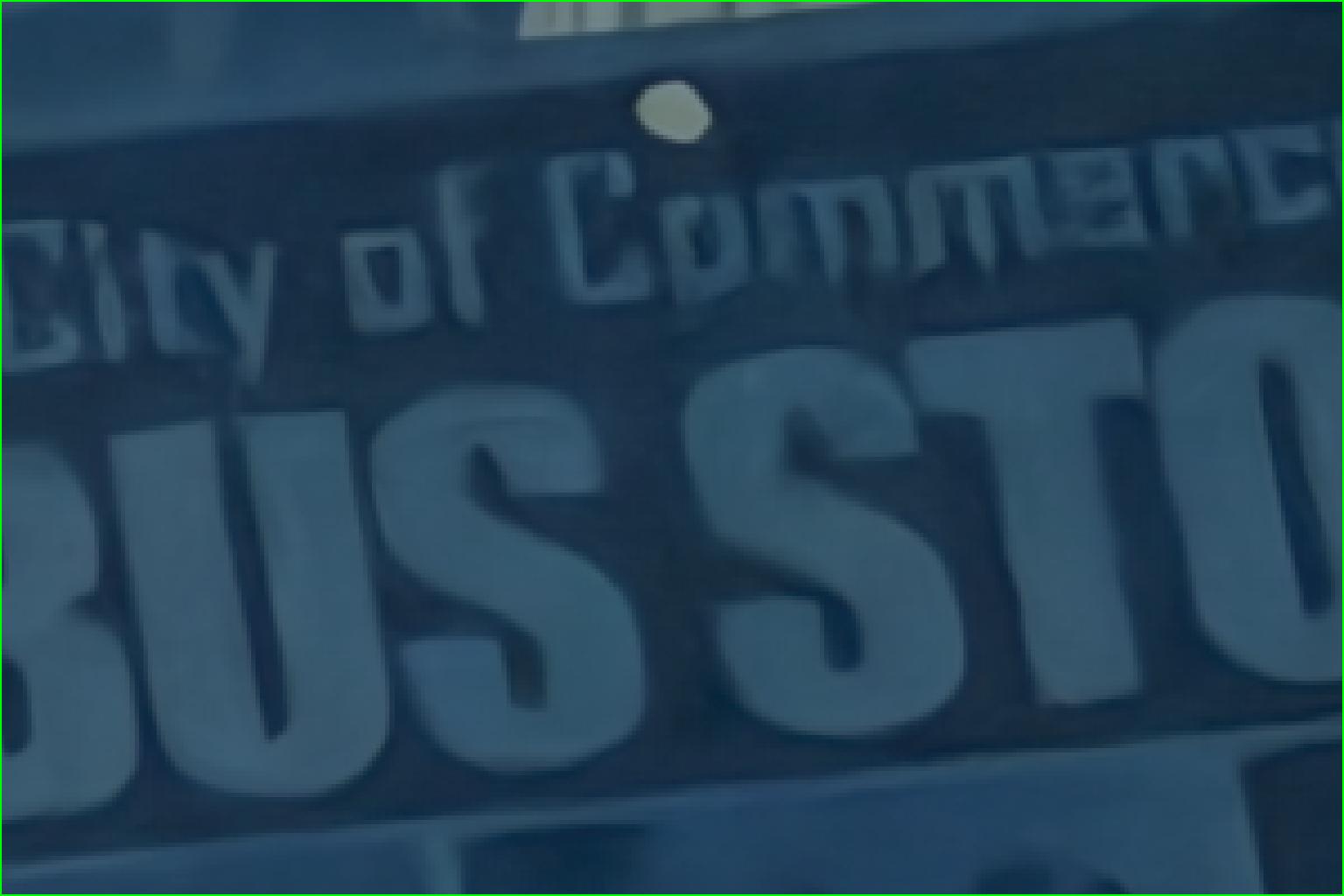}
    {\small (b) EDVR*}
    \end{minipage} 
    \begin{minipage}[b]{0.16\textwidth}
    \centering
    \includegraphics[width=1\textwidth]{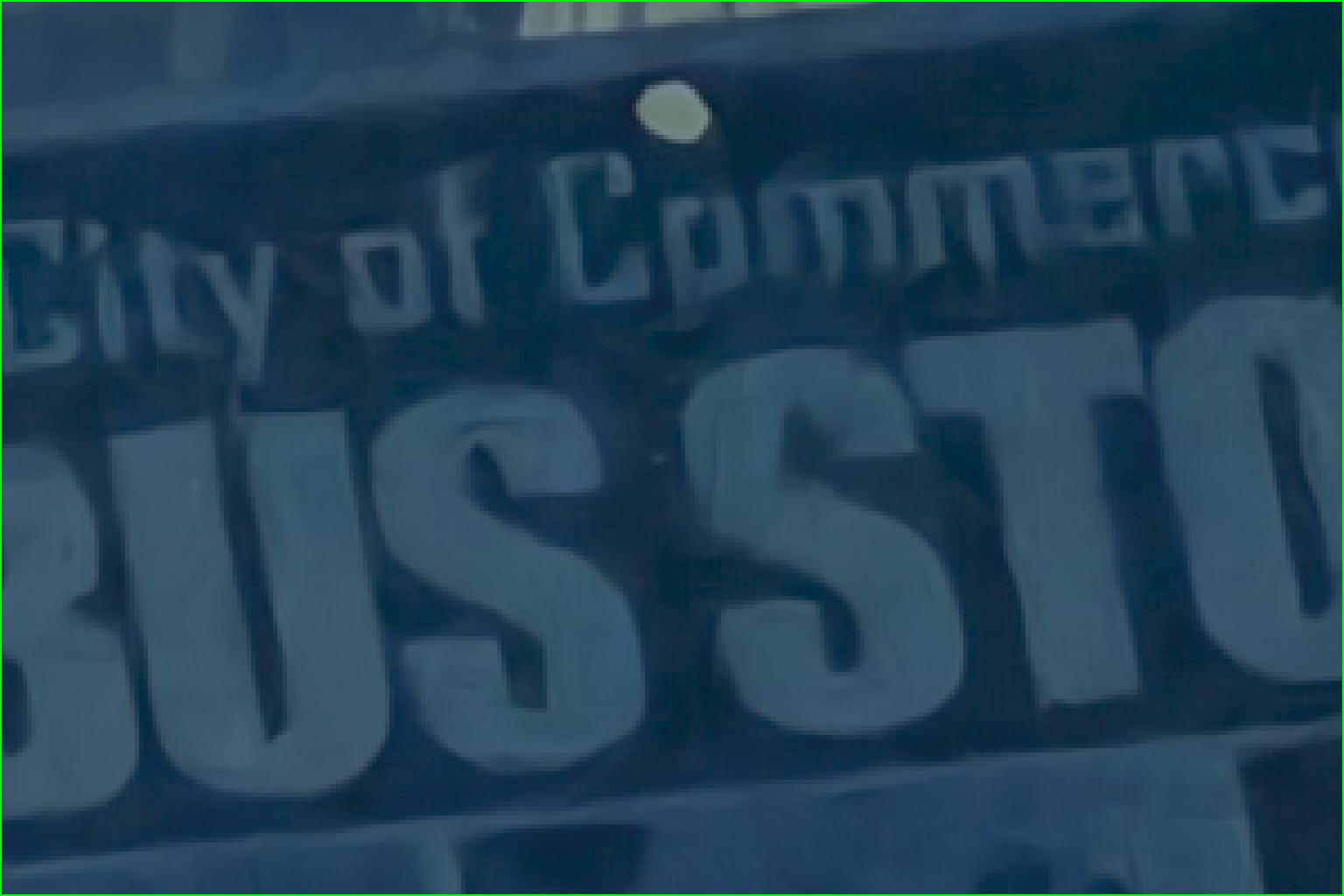}
    {\small (c) RviDeNet*}
    \end{minipage} 
    \begin{minipage}[b]{0.16\textwidth}
    \centering
    \includegraphics[width=1\textwidth]{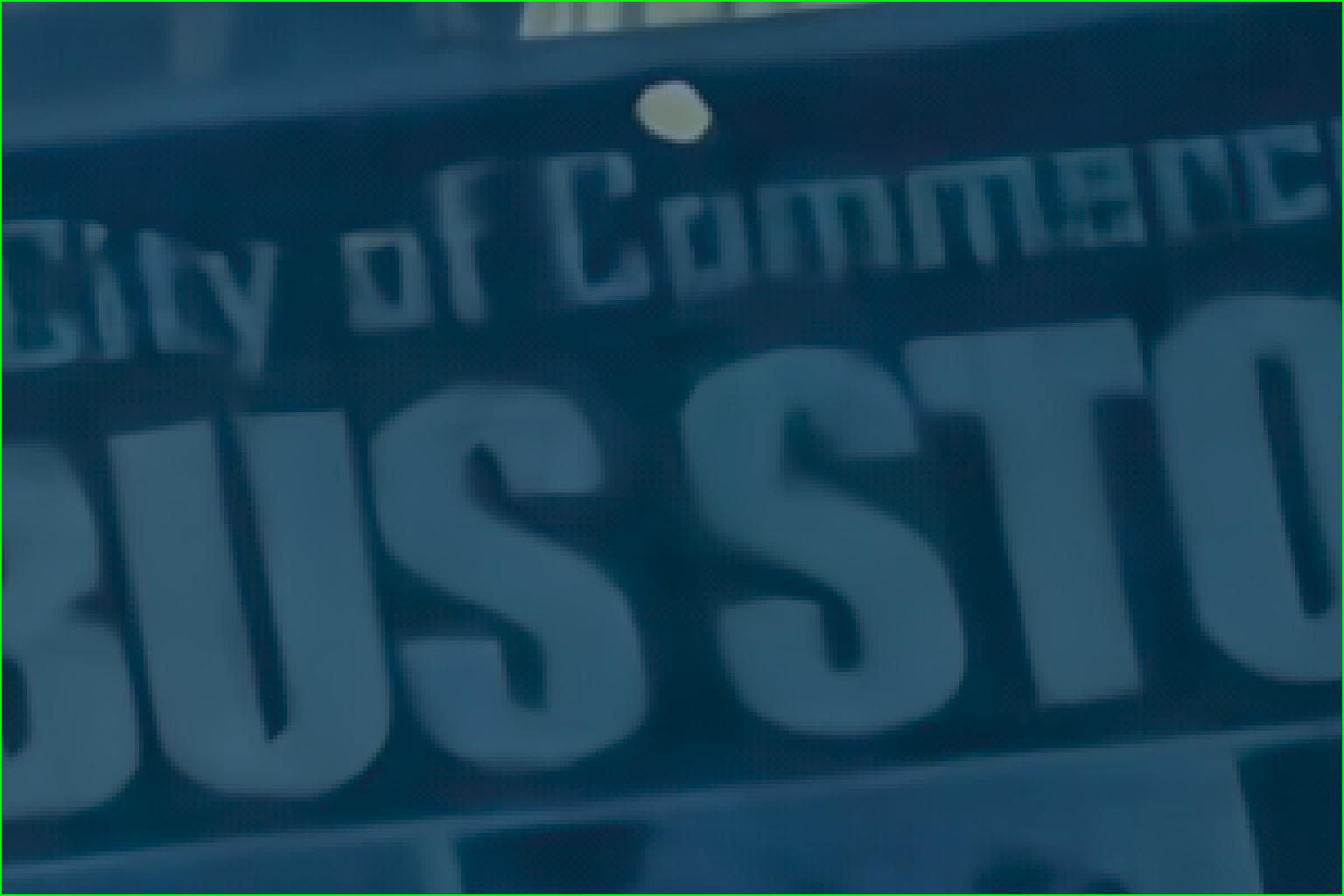}
    {\small (d) GCP-Net}
    \end{minipage} 
    \begin{minipage}[b]{0.16\textwidth}
    \centering
    \includegraphics[width=1\textwidth]{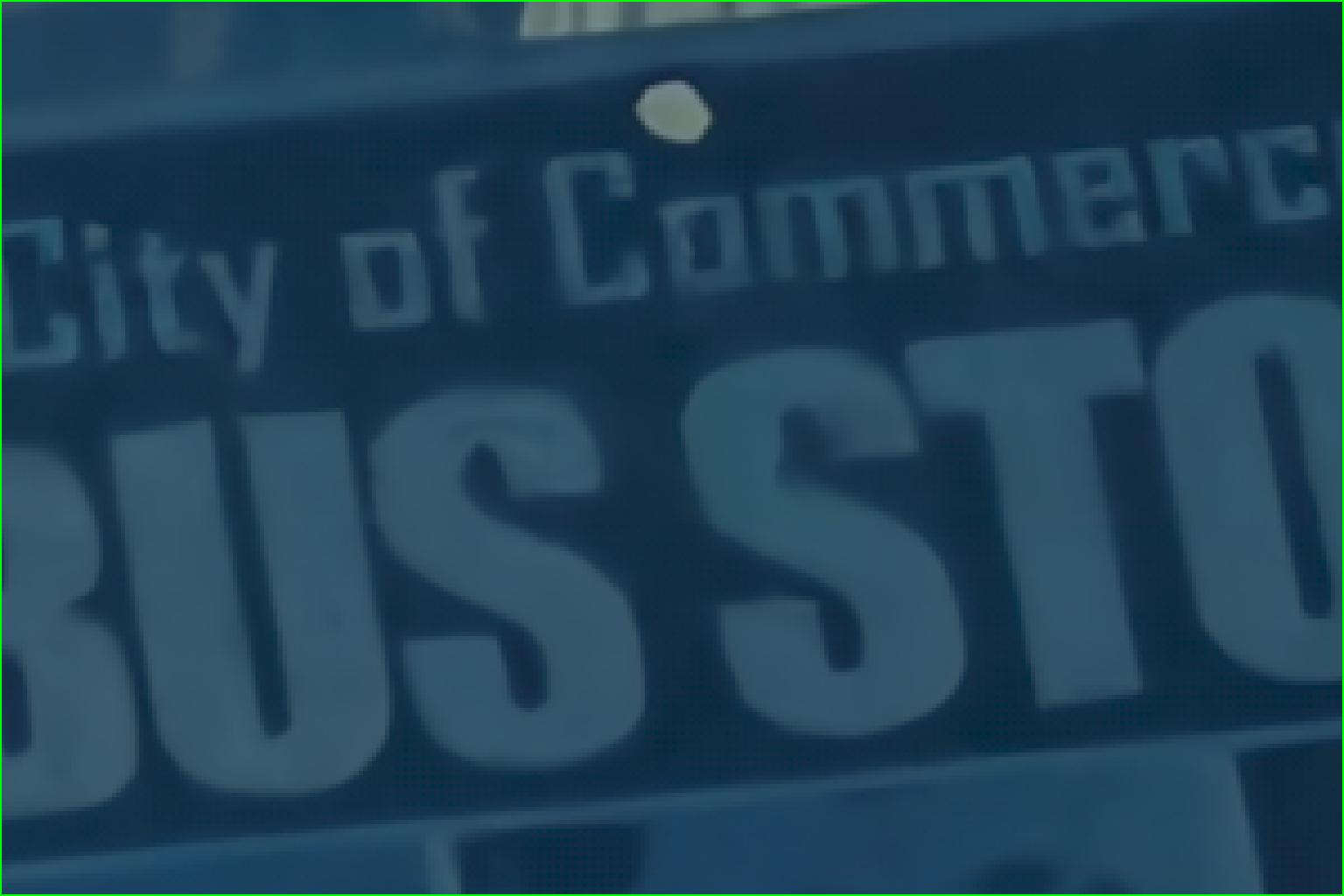}
    {\small (e) Ours}
    \end{minipage} 
    \begin{minipage}[b]{0.16\textwidth}
    \centering
    \includegraphics[width=1\textwidth]{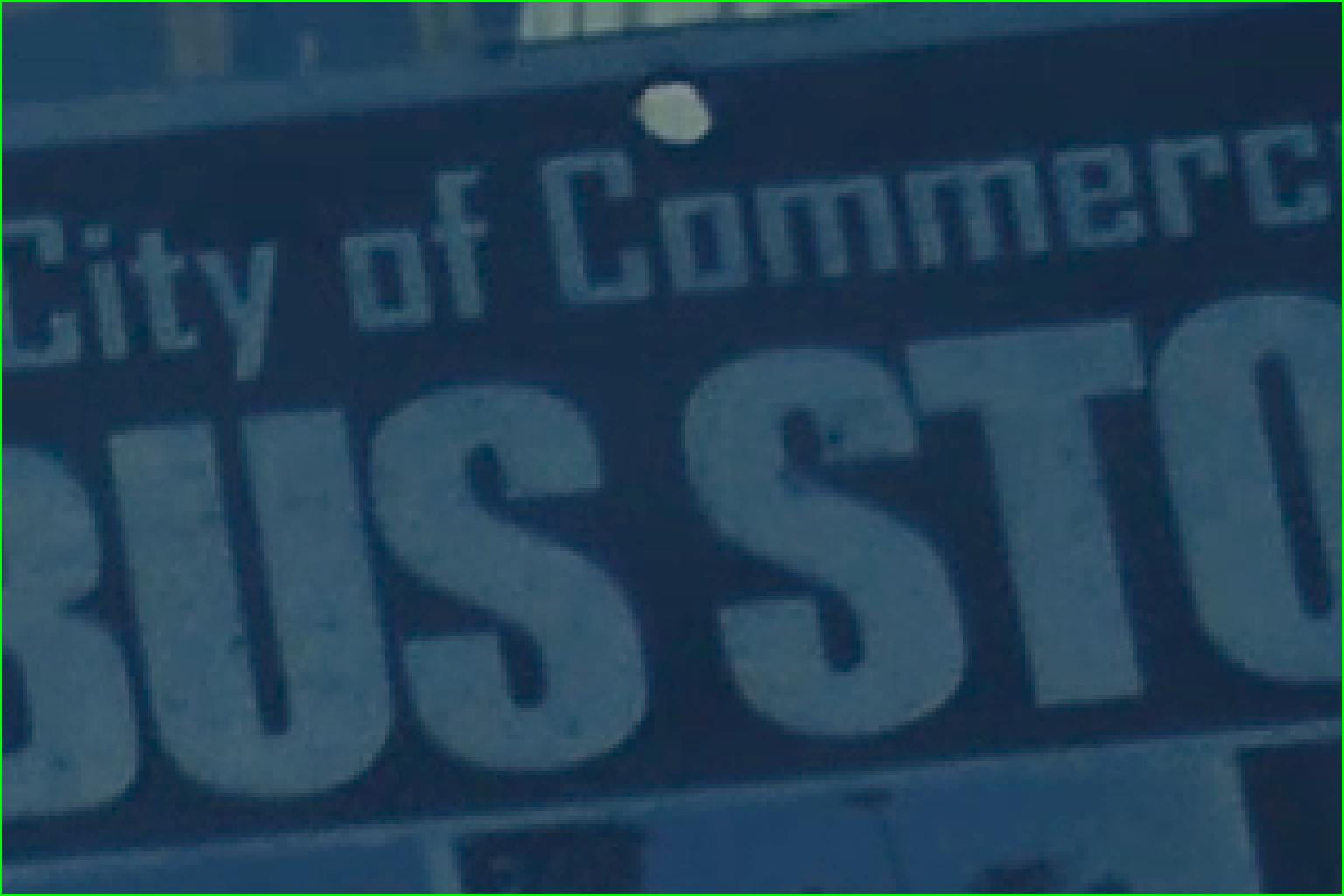}
    {\small (f) GT}
    \end{minipage} 
\end{minipage}
\vspace{0.15cm}

\caption{JDD results of different methods on the Videezy4K dataset.}
\label{fig4K}
\vspace{-0.2cm}
\end{figure*}

\subsection{Design of Training Loss}
\label{DPBMLoss}
\noindent\textbf{DPBM Loss.} Firstly, to avoid learning an all zeros vector or a uniform vector of $w$ in Equ.~\ref{gumbel}, we use penalty when $w$ is not a one-hot vector by calculating the statistical characteristics of $w$, which is denoted as $\mathcal{L}_{one-hot}$. To be specific, the sum and variance of $w$ should equal to 1 and $1/M$, respectively. $M$ is the patch number of query $d(P_{t,i},P_r)_{i\in I}$. The $\mathcal{L}_{one-hot}$ can be written as:  
\begin{equation}
    \mathcal{L}_{one-hot} = \vert \text{sum}(w) - 1 \vert + \vert \text{var}(w) - 1/M \vert,
\end{equation}
where $\text{sum}(w)$ and $\text{var}(w)$ are the sum and variance of $w$.

Due to the corruption of noise in raw images, the BM distances between noisy patches are not accurate. To stabilize the DPBM training process and relieve the impact of noise, we use the BM results obtained by clean images to guide the training process. We let  
\begin{equation}
    \mathcal{L}_{BM} = \Vert d(P_{t,i},P_r) - d(P_{t,i}^{c},P_r^{c}) \Vert_2^2,
\end{equation}
where $P_{t,i}^{c}$ and $P_r^{c}$ are the query patches in frame $t$ and target patch in frame $r$ using clean images. $\mathcal{L}_{BM}$ is utilized in the first 20W iterations.

\noindent\textbf{Interpolation Loss.} 
We design an interpolation loss, denoted by $\mathcal{L}_{ip}$, to encourage the network to make better use of the information from other frames.
The core idea of $\mathcal{L}_{ip}$ is to use neighboring frames to ``interpolate" the reference frame. 
In the fusion module, we aggregate the features from all other frames, excluding the reference frames, to interpolate the reference frame using the reconstruction UNet. The ``interpolation" output is denoted as $\hat{x}^i$. To encourage the aligned features to capture more textures, we calculate $\mathcal{L}_{ip}$ only on high frequency regions (denoted by $m_h$), which can be obtained using ~\cite{liu2020joint}.
The $\mathcal{L}_{ip}$ can be written as:
\begin{equation}
    \mathcal{L}_{ip} = m_h\odot\sqrt{\Vert \hat{x}^i - x \Vert^2 + \epsilon^2},
\end{equation}
where $\sqrt{\Vert \hat{x} - x \Vert^2 + \epsilon^2}$ is the Charbonnier penalty function, and $\epsilon$ is set to 0.001. 

\noindent\textbf{Overall Loss.}
Denote by $\hat{x}$ the reconstructed clean full-color image from the noisy raw burst data and by $x$ the ground-truth clean image of the reference frame. A reconstruction loss can be defined as:
\begin{equation}
    \mathcal{L}_{r} = \sqrt{\Vert \hat{x} - x \Vert^2 + \epsilon^2} + \sqrt{\Vert \Gamma(\hat{x}) - \Gamma(x) \Vert^2 + \epsilon^2},
\end{equation}
in which $\Gamma(\cdot)$ is the ISP operator consisting of white balance, color correction and gamma compression. The implementation details of $\Gamma(\cdot)$ can be found in \cite{brooks2019unprocessing}.

Overall, the total loss of our model is:
\begin{equation}
    \mathcal{L} = \mathcal{L}_{r} + \beta\mathcal{L}_{ip} + \rho\mathcal{L}_{one-hot} + \eta\mathcal{L}_{BM},
\end{equation}
where $\beta$ and $\eta$ are the balancing parameters and are set to 1 and $1\times 10^3$. The penalty factor $\rho$ is set to $1\times 10^5$. 

\begin{figure*}[!t]
\setlength{\abovecaptionskip}{0.0cm}
\setlength{\belowcaptionskip}{-0.cm}
\centering
\subfloat{
\begin{minipage}[t]{0.24\textwidth}
\centering
\includegraphics[width=1\textwidth]{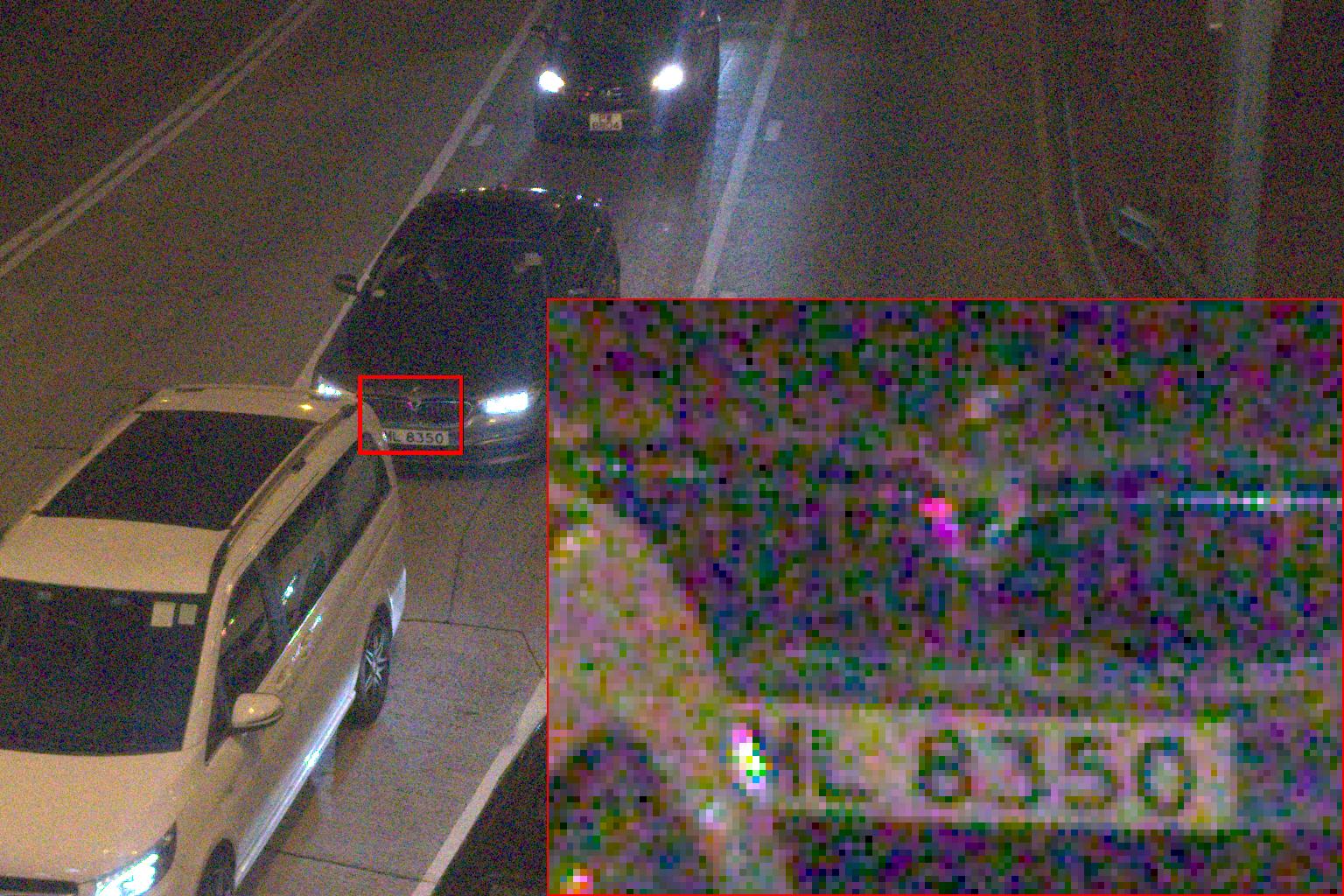}
{\footnotesize  (a) Noisy image}
\end{minipage} 
\begin{minipage}[t]{0.24\textwidth}
\centering
\includegraphics[width=1\textwidth]{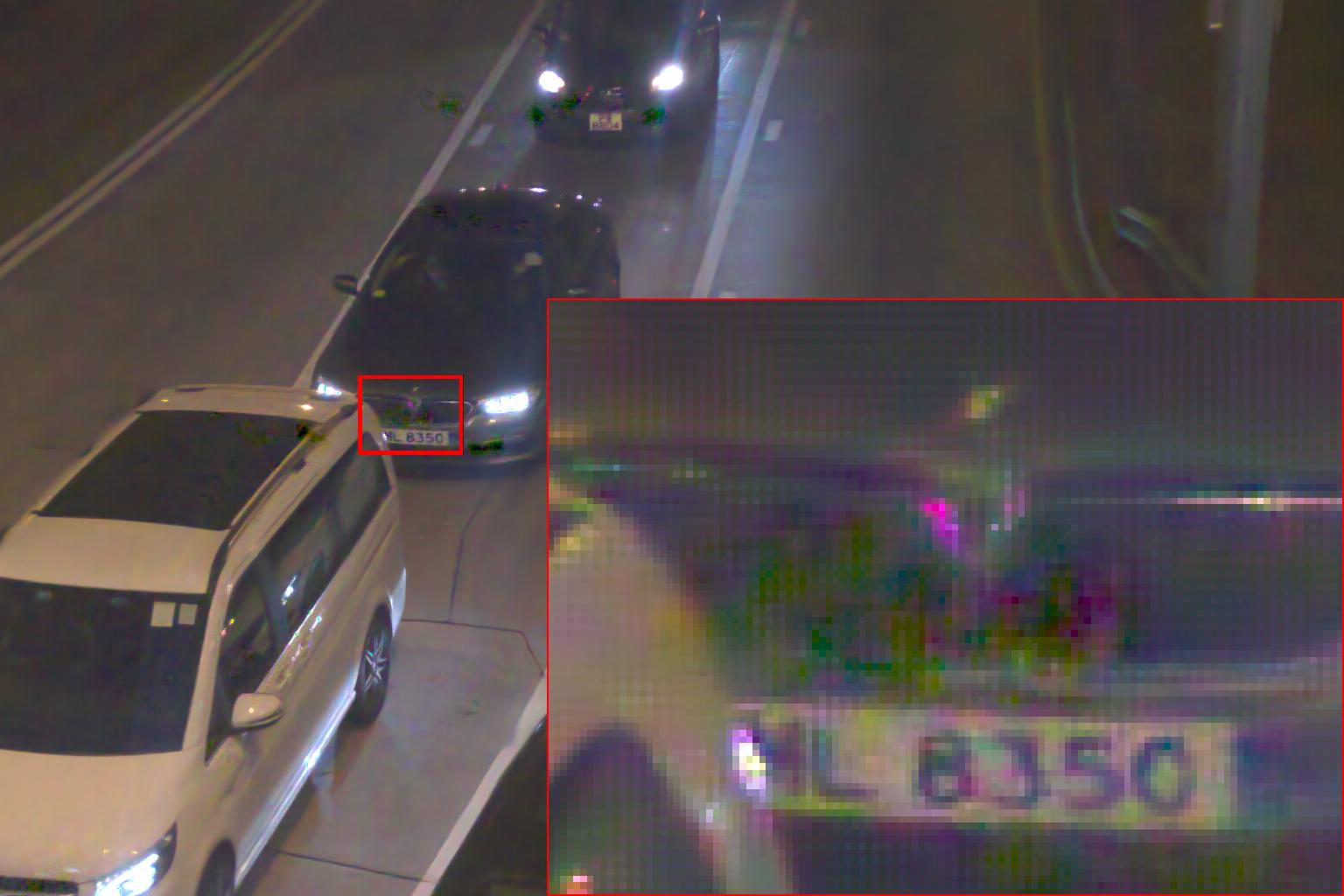}
{\footnotesize  (b) KPN+DMN}
\end{minipage} 
\begin{minipage}[t]{0.24\textwidth}
\centering
\includegraphics[width=1\textwidth]{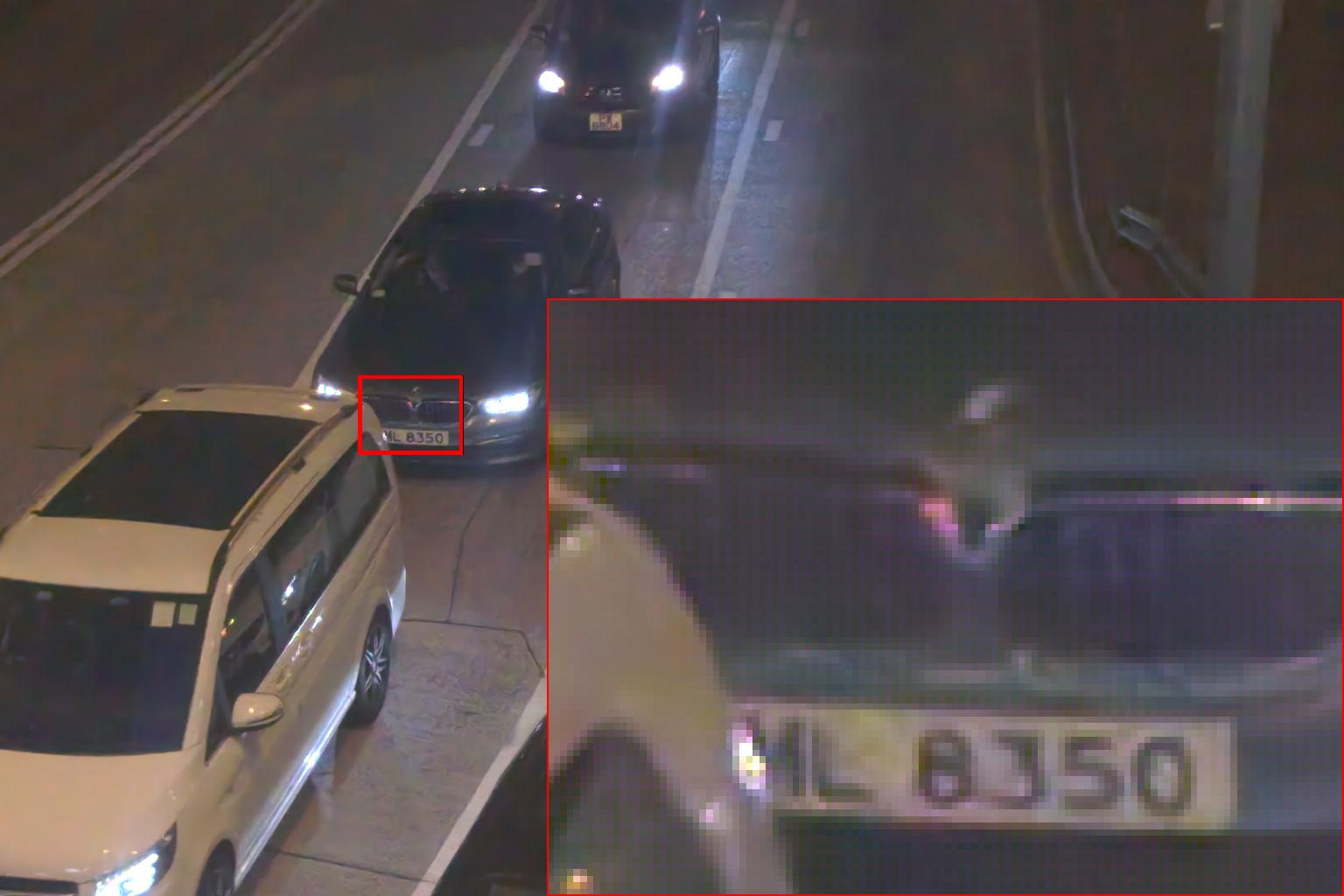}
{\footnotesize  (c) EDVR+DMN}
\end{minipage} 
\begin{minipage}[t]{0.24\textwidth}
\centering
\includegraphics[width=1\textwidth]{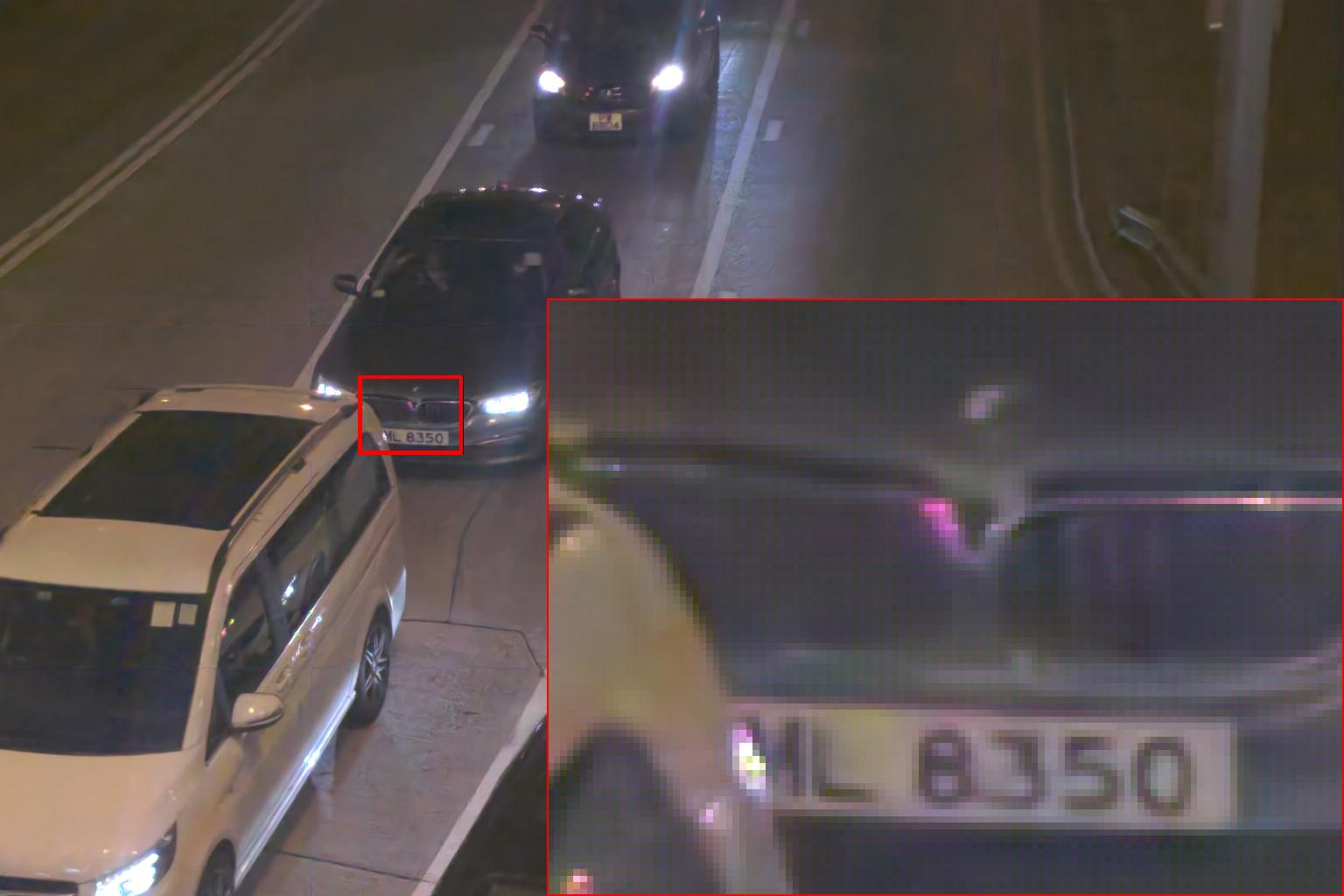}
{\footnotesize  (d) RviDeNet+DMN}
\end{minipage} 
}

\subfloat{
\begin{minipage}[t]{0.24\textwidth}
\centering
\includegraphics[width=1\textwidth]{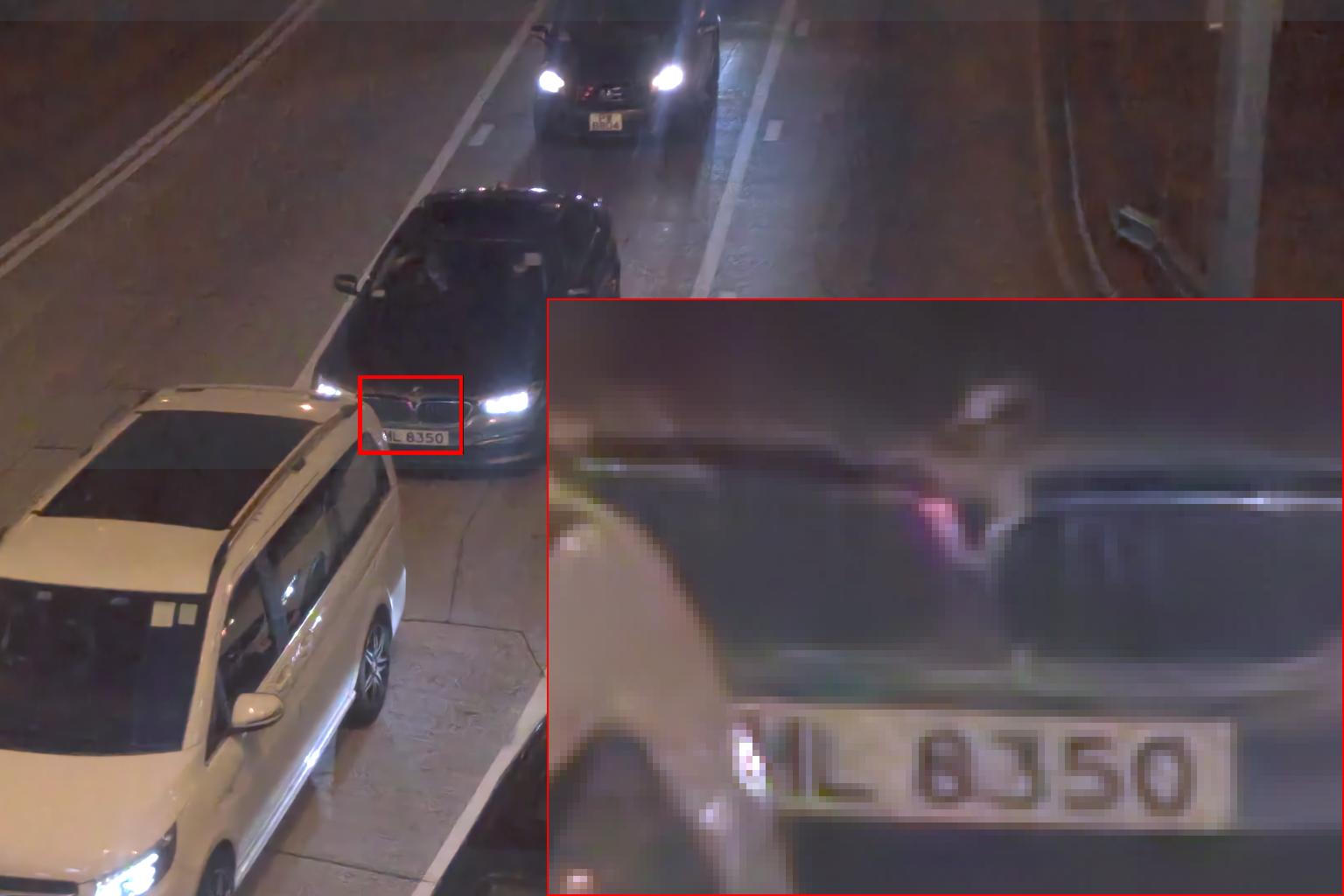}
{\footnotesize  (e) EDVR*}
\end{minipage} 
\begin{minipage}[t]{0.24\textwidth}
\centering
\includegraphics[width=1\textwidth]{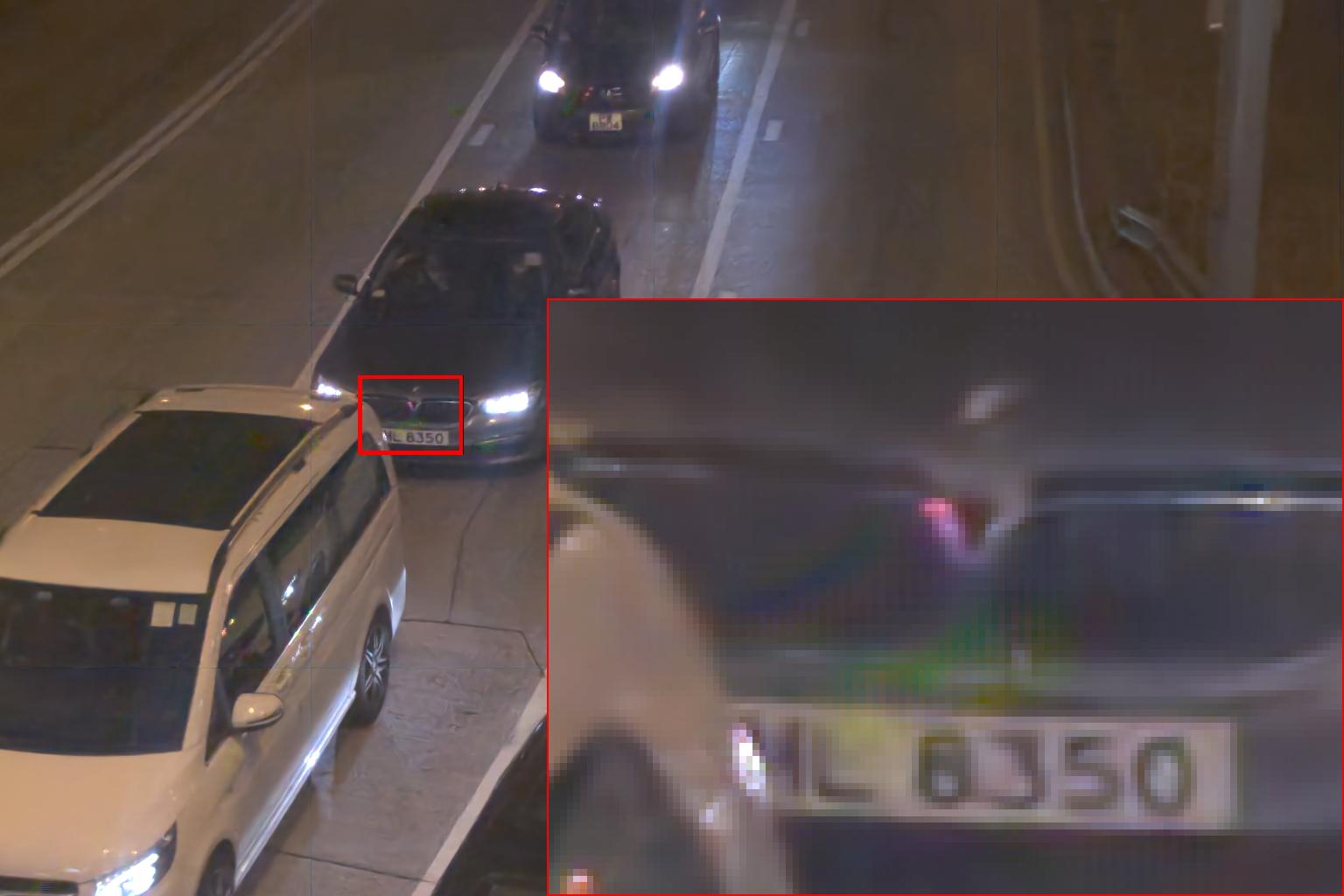}
{\footnotesize  (f) RviDeNet*}
\end{minipage} 
\begin{minipage}[t]{0.24\textwidth}
\centering
\includegraphics[width=1\textwidth]{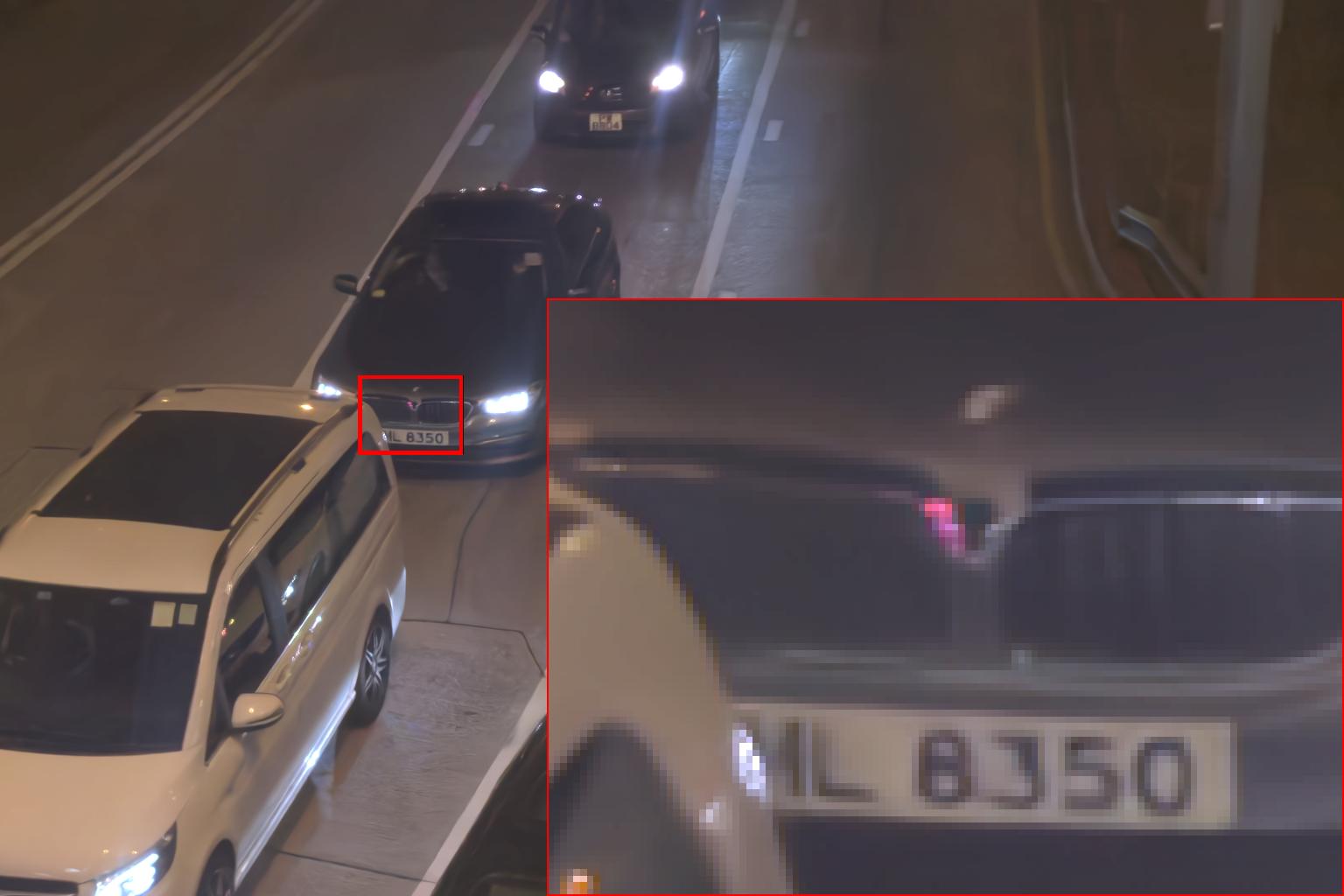}
{\footnotesize  (g) GCP-Net}
\end{minipage} 
\begin{minipage}[t]{0.24\textwidth}
\centering
\includegraphics[width=1\textwidth]{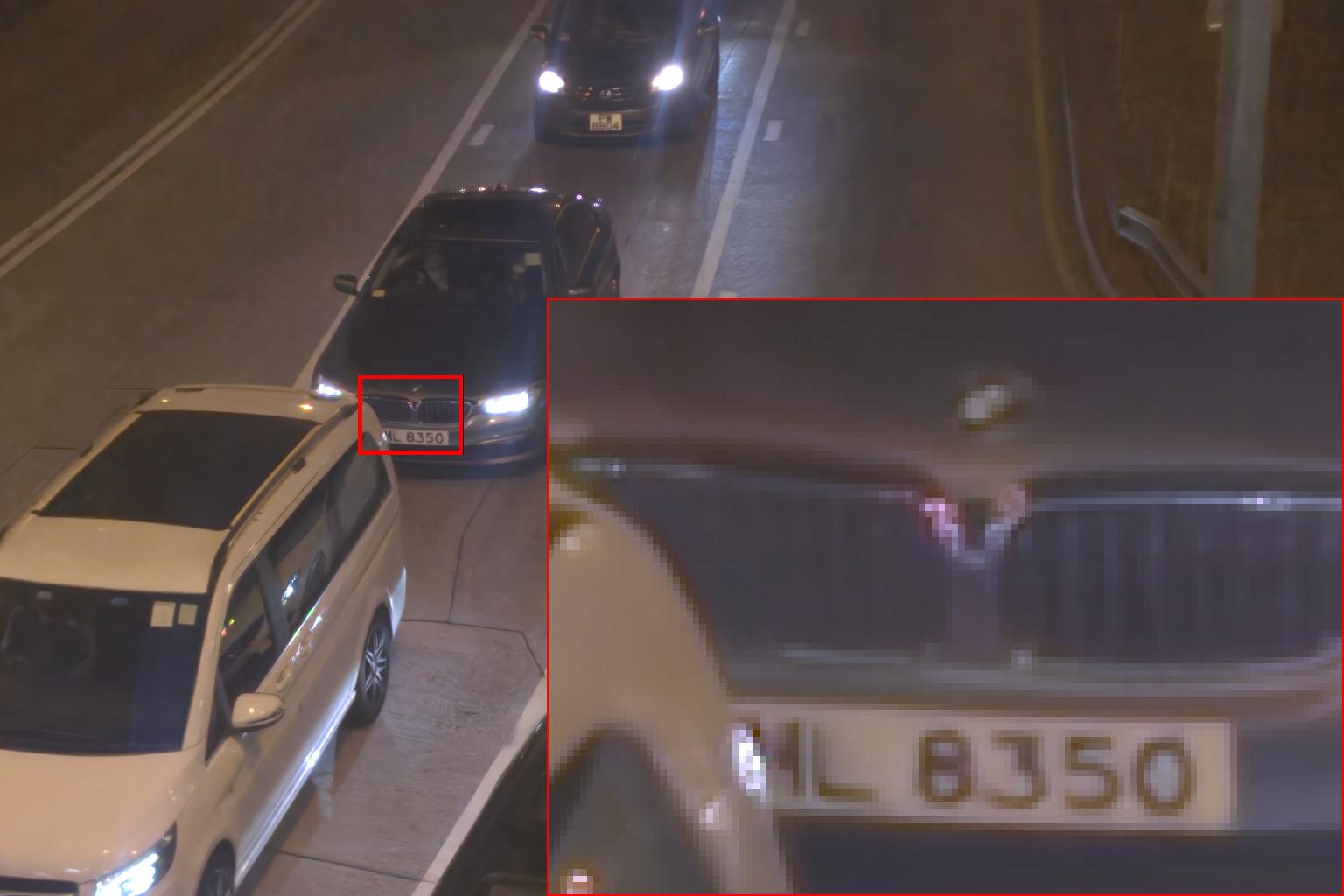}
{\footnotesize  (h) Ours}
\end{minipage} 
}
\caption{JDD-B results on real-world burst images by different methods.}
\label{figReal}
\vspace{-0.4cm}
\end{figure*}

\section{Experiments}
\label{sec:exp}
\subsection{Model Training Details}
\label{sec:expset}
\noindent\textbf{Training Data.} We follow the method in~\cite{brooks2019unprocessing} to synthesize training data for real-world burst images. The 240 training clips of 720p in the REDS dataset~\cite{nah2019ntire} are used to generate training data. Firstly, the sRGB videos are unprocessed to linear RGB space using~\cite{brooks2019unprocessing}. The obtained sequences are taken as the clean ground-truth images $x_t$. 
The inputs of the network $\{y_t\}_{t=1}^{N}$ are the noisy raw burst images which are obtained by~\cite{foi2008practical}:
\begin{equation}
    y_t = M(x_t)+n(M(x_t),\sigma_s,\sigma_r),
\label{noisemodel}
\end{equation}
where $n(x, \sigma_s, \sigma_r) \sim \mathcal{N}(0, \sigma_s x + \sigma_r^2)$, and $M(\cdot)$ is the mosaic downsampling operator. $\sigma_s$ and $\sigma_r$ represent the scale of shot noise and read noise, respectively.  

\noindent\textbf{Training Details.} We use the center frame as the reference frame. Following \cite{mildenhall2018burst,guo2021joint}, the $\sigma_s$ and $\sigma_r$ are uniformly sampled in the ranges of $[10^{-4}, 10^{-2}]$ and $[10^{-3}, 10^{-1.5}]$. Our model is implemented in PyTorch and is trained using two RTX 2080Ti GPUs. During training, we use the Adam~\cite{kingma2014adam} optimizer with momentum 0.9. The learning rates are initialized as $1\times 10^{-5}$ for the lightweight network in coarse alignment and $1\times 10^{-4}$ for other network parts. Then the learning rate is decreased using the cosine function~\cite{loshchilov2016sgdr}.

\subsection{Comparison with State-of-the-Art Methods}
\noindent\textbf{Competing Methods.} We compare our method with the recently developed JDD-B method GCP-Net~\cite{guo2021joint}. Following~\cite{guo2021joint}, we also combine several SOTA burst denoising algorithms, \ie, KPN~\cite{mildenhall2018burst}, EDVR~\cite{wang2019edvr} and RviDeNet~\cite{yue2020supervised}, with a SOTA demosiacking method DMN~\cite{gharbi2016deep} for comparison. EDVR~\cite{wang2019edvr} and RviDeNet~\cite{yue2020supervised} are also slightly modified to adapt to the JDD-B task by adding an upsampling operator to the output layer, and we denote them as EDVR* and RviDeNet*. All competing models are retrained using our experiment setting. 

\noindent\textbf{Results on Synthetic Data.} To quantitatively evaluate the competing methods, we first perform experiments on two synthetic datasets, \emph{i.e.}, REDS4~\cite{wang2019edvr} and a dataset collected by us, called Videezy4K. REDS4 is widely used as the test set in the study of video super-resolution, whose resolution is 720p. Since images/videos captured by modern smartphone cameras and DSLRs commonly have 4K resolution, the shift between frames can be larger than the REDS4 dataset. To better evaluate the performance on 4K videos, we collected 12 clips of 4K videos from the Videezy website~\cite{videezy_dataset} and choose 20 frames of each video for testing. All testing videos are firstly converted to linear RGB space using the same pipeline as generating training data and the noisy burst raw images are generated by adding heteroscedastic Gaussian noise with Equ.~\ref{noisemodel}. 

The quantitative results on REDS4 and Videezy4K datasets are shown in Tables~\ref{synthREDS4} and \ref{synthVideezy}, respectively. Following~\cite{mildenhall2018burst,guo2021joint}, we calculate the PSNR and SSIM indices after gamma correction to better reflect perceptual quality. The qualitative comparisons on REDS4 and Videezy4K are presented in Figs.~\ref{figRED} and \ref{fig4K}, respectively. 
We can see that due to the correlation between denoising and demosaicking, JDD-B algorithms (\ie, EDVR* and RviDeNet*) obtain better performance than EDVR+DMN and RviDeNet+DMN. However, limited by the small receptive field of one-stage alignment module, EDVR*, RviDeNet* and GCP-Net suffer from the over-smoothing problem in areas with large motion. Benefiting from the two-stage alignment, our model recovers more textures, especially on the moving objects. In Fig.~\ref{figRED}, the horizontal stripes on the walking people are well reconstructed with little zippers and color artifacts by our model. In Fig.~\ref{fig4K}, the characters can be more clearly restored and identified by our method. For images with small motion, \ie, \emph{Clip000} in the REDS4 dataset, our method still obtains better performance. More visual results can be found in the supplementary file.

\begin{figure*}[!t]
\setlength{\abovecaptionskip}{0.1cm}
\setlength{\belowcaptionskip}{-0.0cm}
\centering
\begin{minipage}[b]{0.265\textwidth}
\centering
    \begin{minipage}[b]{1.0\textwidth}
    \centering
    \subfloat{
        \includegraphics[width=1\textwidth]{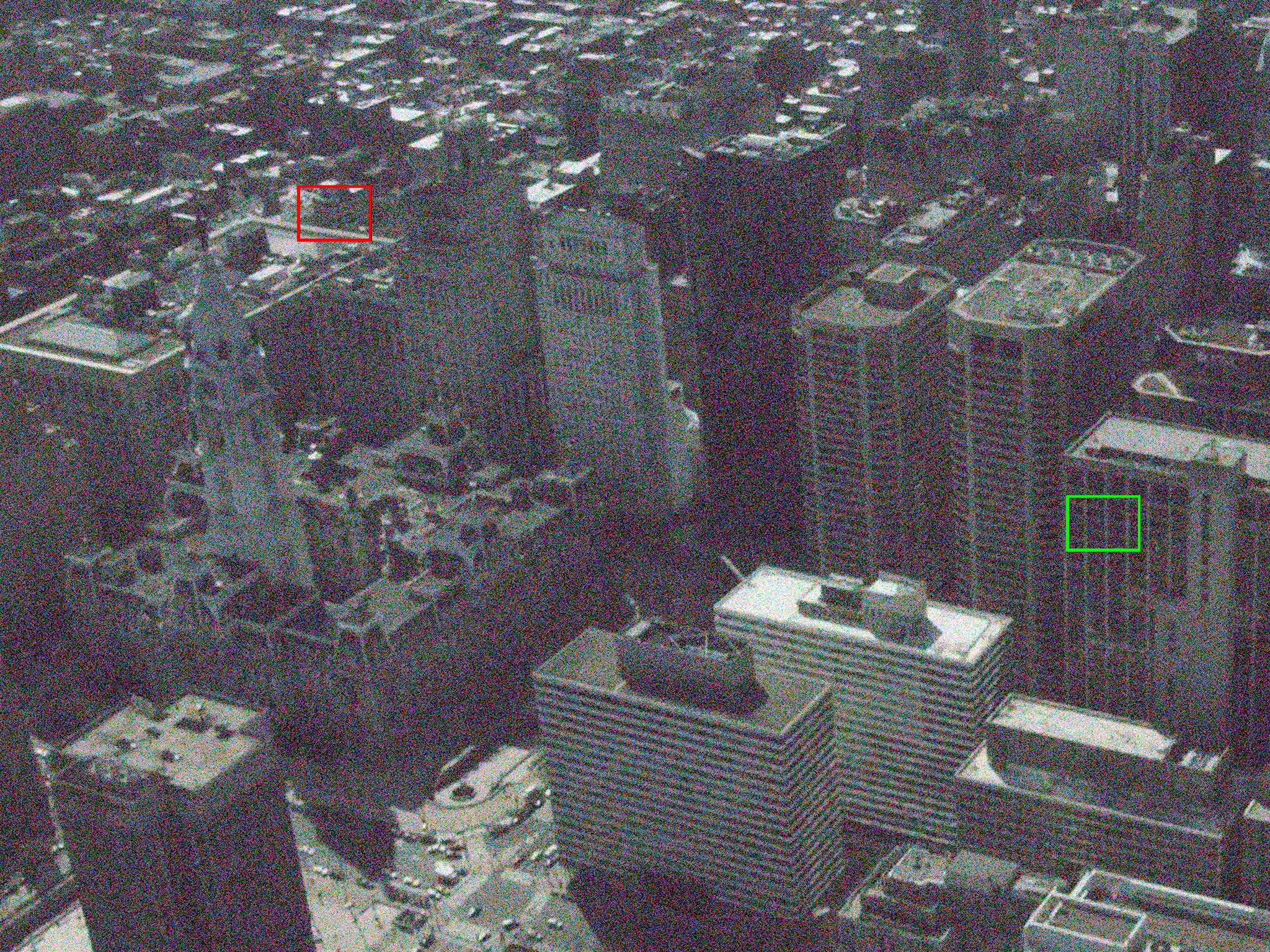}
        }
    \end{minipage}
    {\scriptsize (a) Noisy image}
\end{minipage} \hspace{0.5mm}
\begin{minipage}[b]{0.13\textwidth}
        \centering
        \includegraphics[width=1\textwidth]{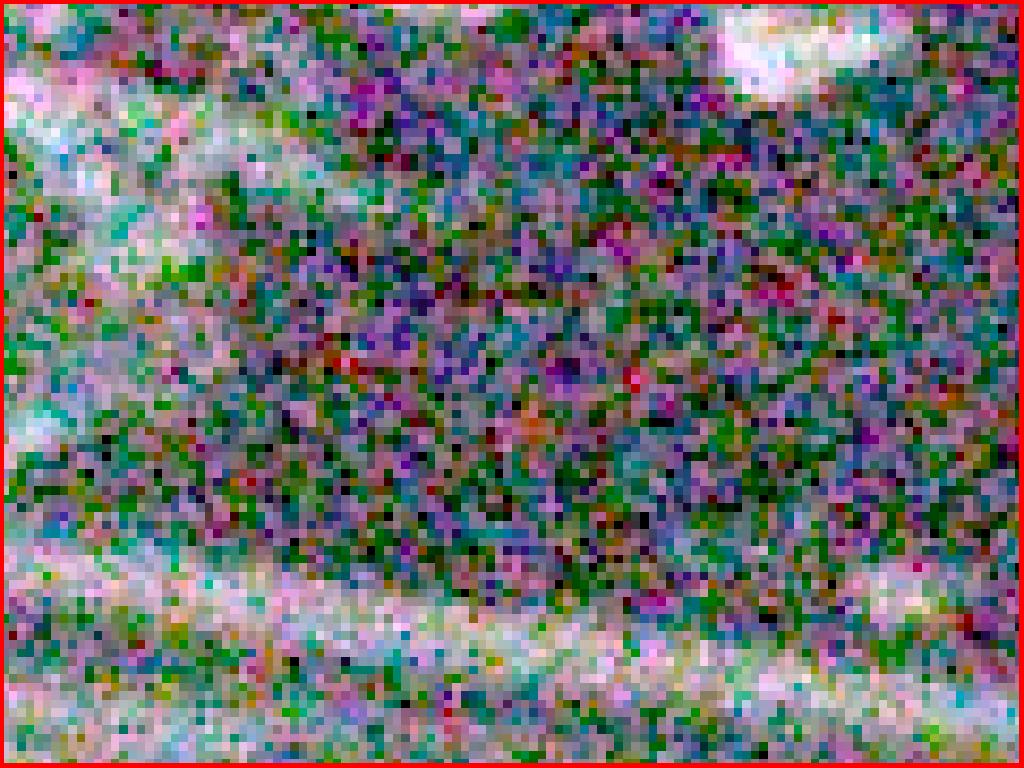}
        
        \vspace{0.4mm}
        \centering 
        \includegraphics[width=1\textwidth]{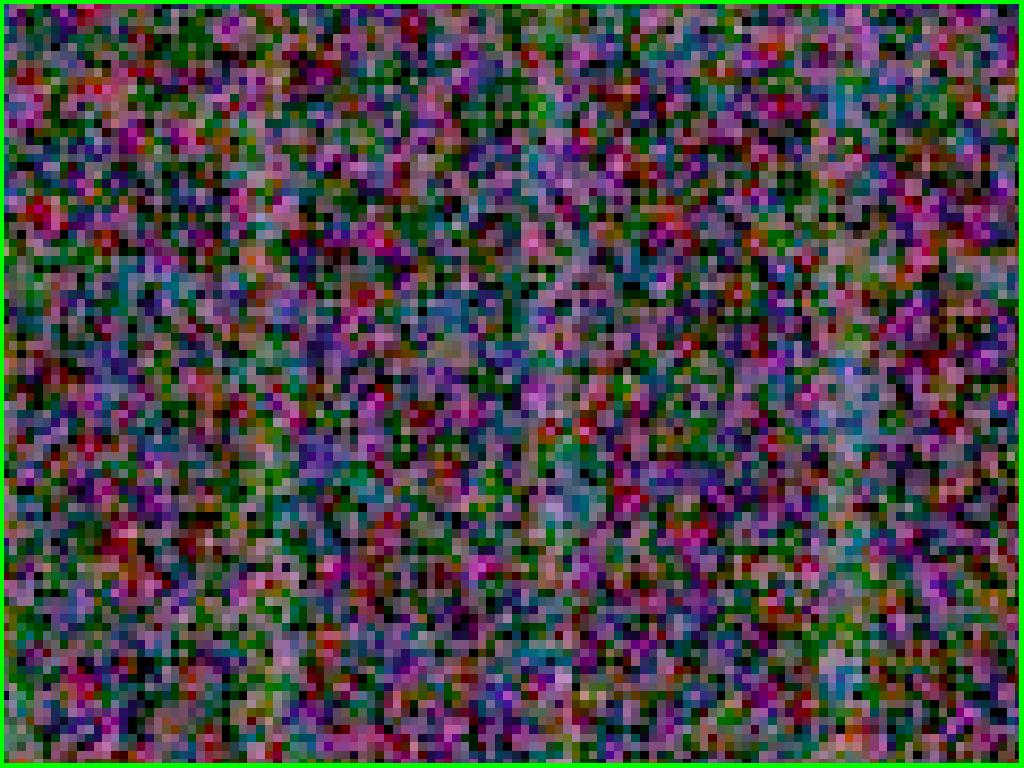}
{\scriptsize (b) Noisy Patch}
\end{minipage} 
\begin{minipage}[b]{0.13\textwidth}
        \centering
        \includegraphics[width=1\textwidth]{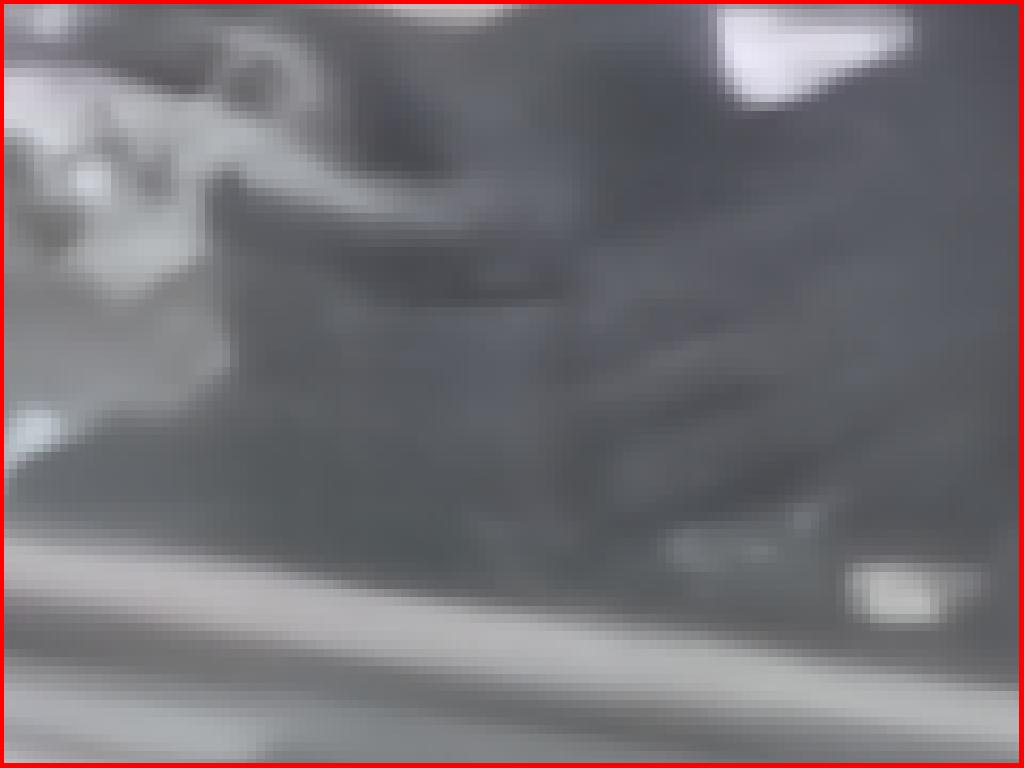}
        
        \vspace{0.4mm}
        \centering
        \includegraphics[width=1\textwidth]{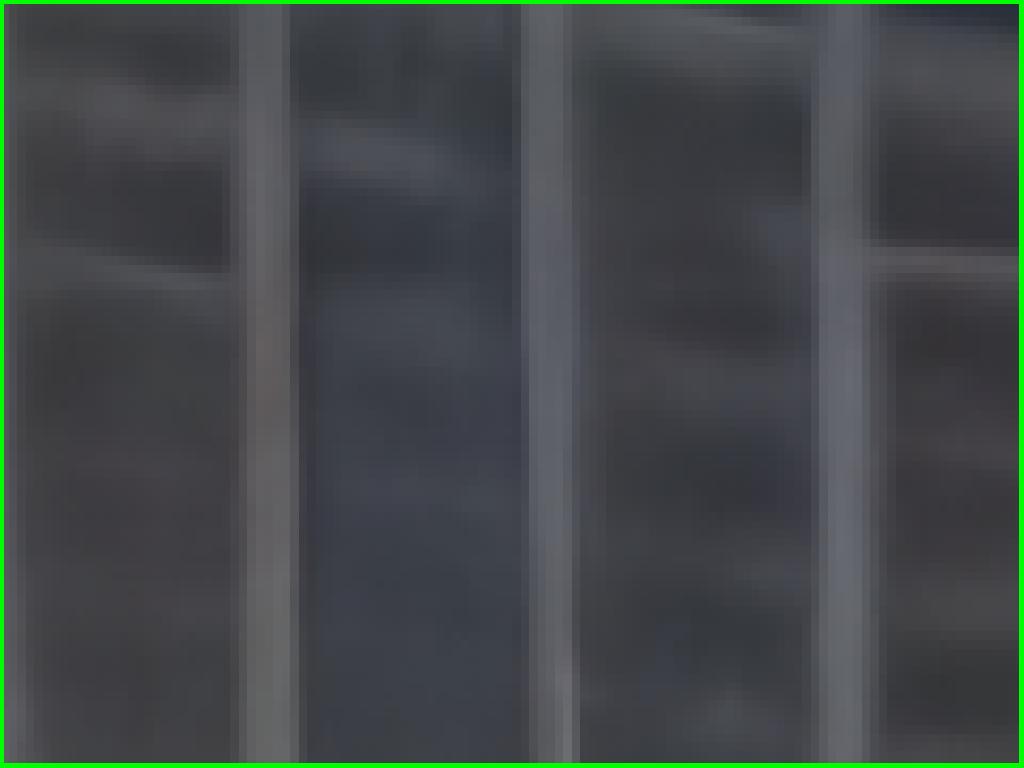}
{\scriptsize (c) w/o Align}
\end{minipage} 
\begin{minipage}[b]{0.13\textwidth}
        \centering
        \includegraphics[width=1\textwidth]{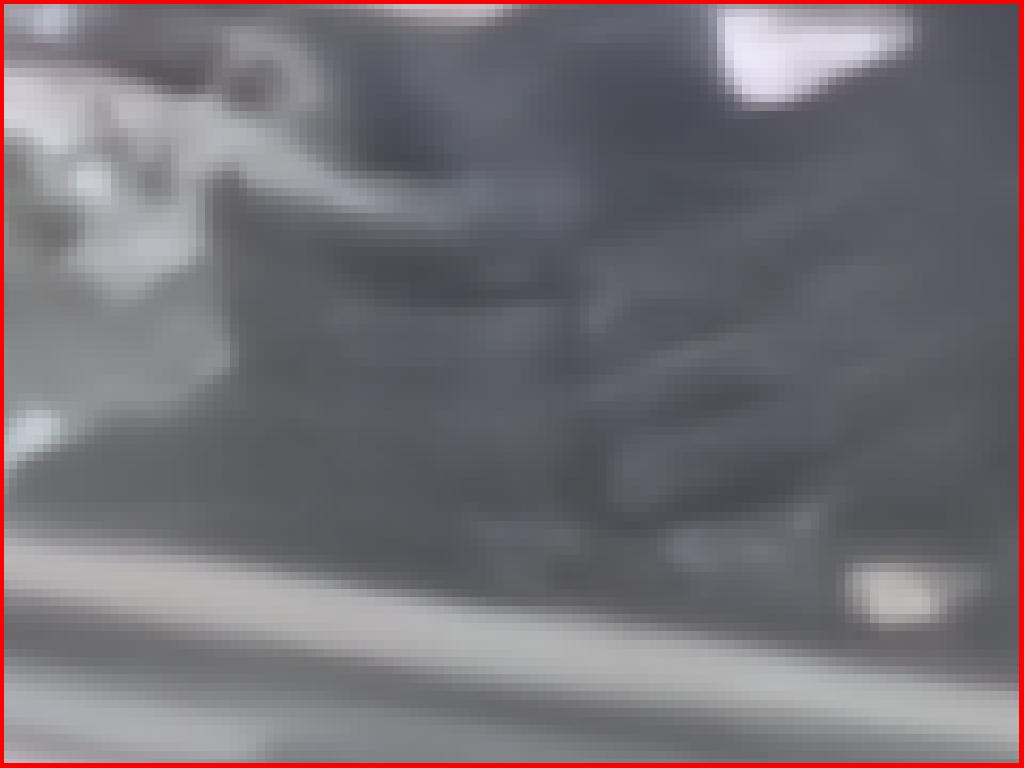}
        
        \vspace{0.4mm}
        \centering
        \includegraphics[width=1\textwidth]{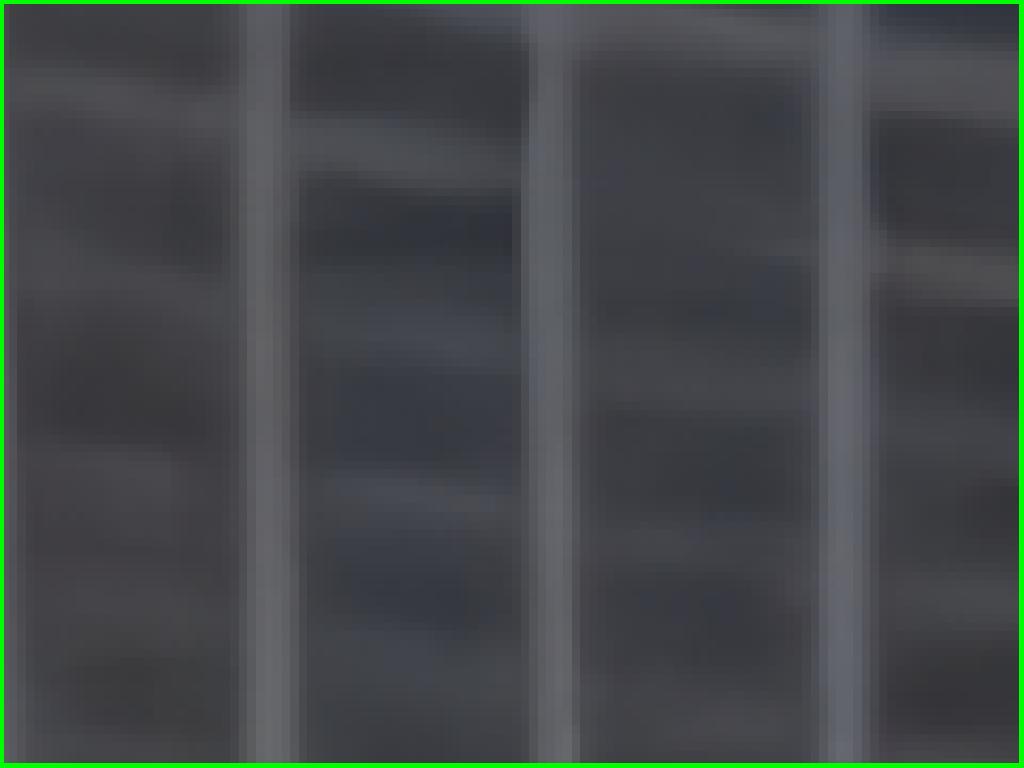}
{\scriptsize (d) w/ RA}
\end{minipage} 
\begin{minipage}[b]{0.13\textwidth}
        \centering
        \includegraphics[width=1\textwidth]{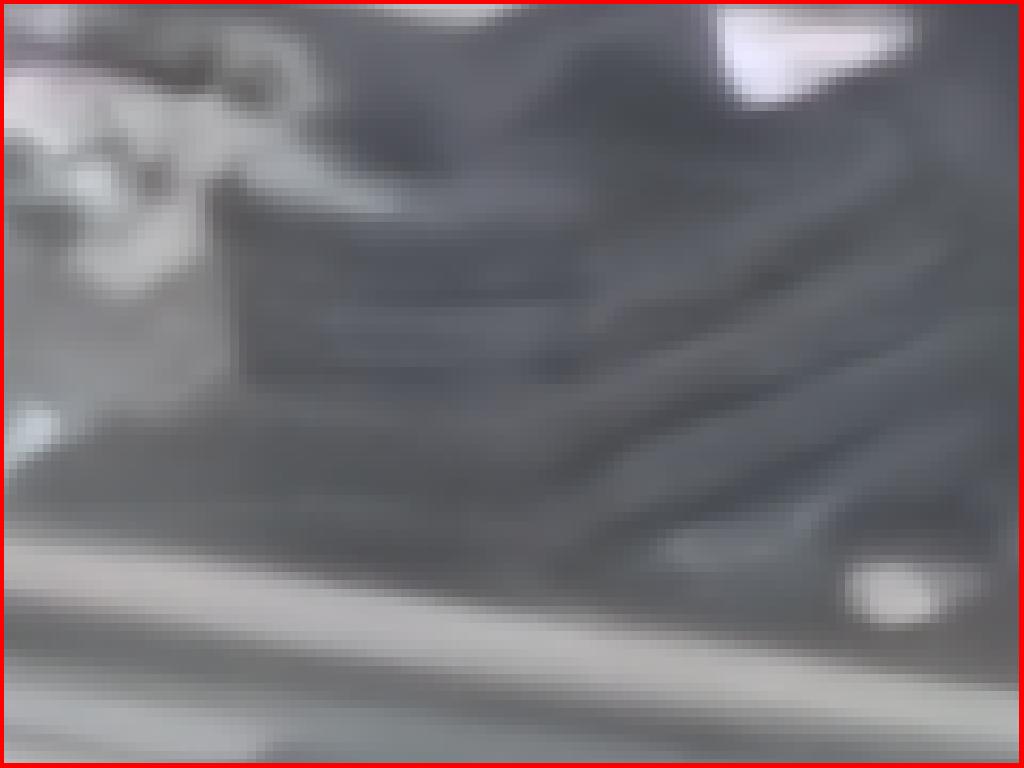}
        
        \vspace{0.4mm}
        \centering
        \includegraphics[width=1\textwidth]{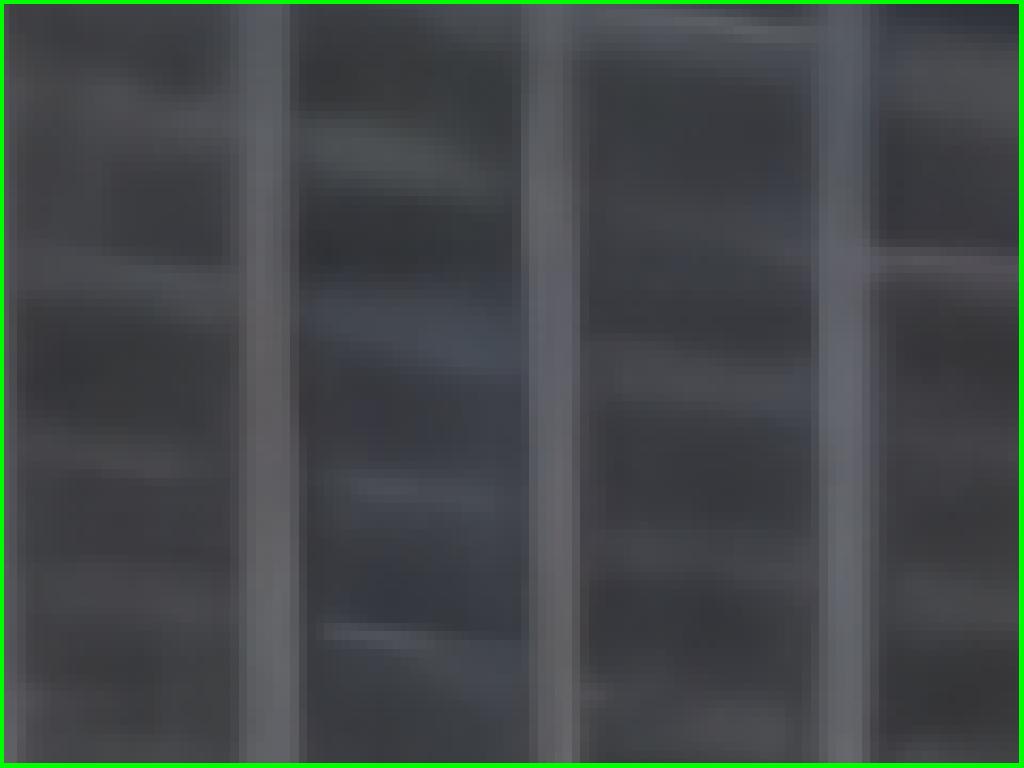}
{\scriptsize (e) w/ CA}
\end{minipage} 
\begin{minipage}[b]{0.13\textwidth}
        \centering
        \includegraphics[width=1\textwidth]{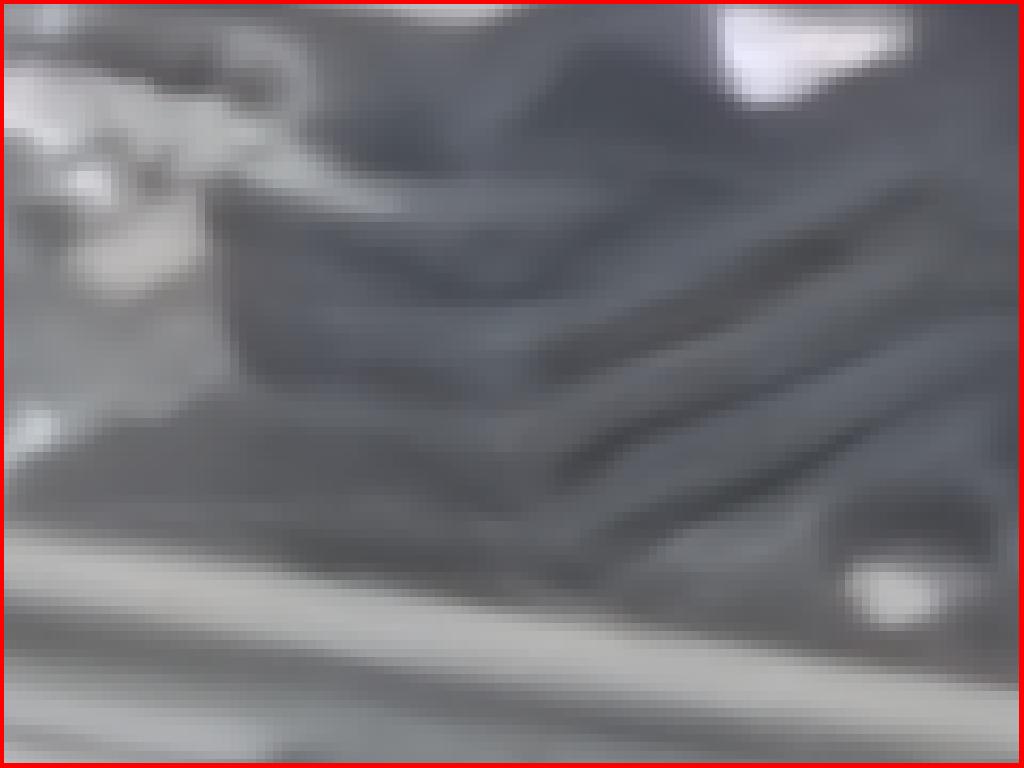}

        \vspace{0.4mm}
        \centering
        \includegraphics[width=1\textwidth]{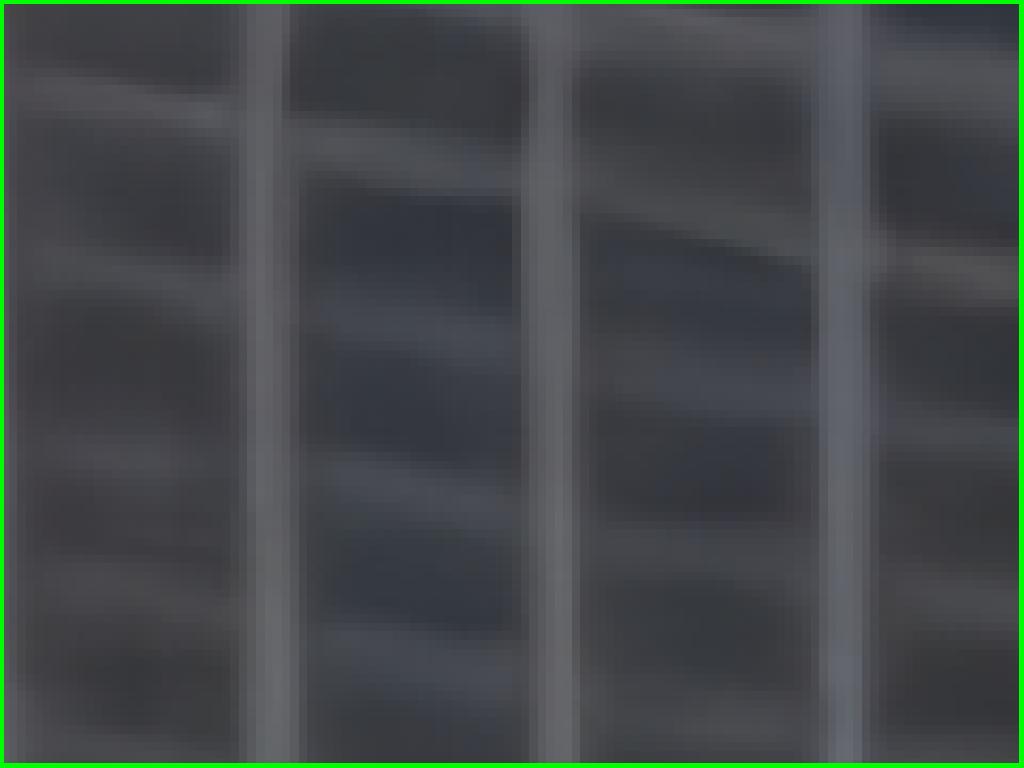}
{\scriptsize (f) two-stage}
\end{minipage} 
\caption{JDD-B results by using different variants of our two-stage alignment model.}
\label{disalign}
\vspace{-0.4cm}
\end{figure*}

\noindent\textbf{Results on Real-world Burst Images.}
We then qualitatively evaluate our method on real-world burst images. The SCBurst dataset~\cite{guo2021joint} is used, where burst images are captured using smartphones with a wide range of ISO values. 
SCBurst contains noisy raw images and the corresponding metadata, which are used in the testing and visualization.
We provide visual comparisons in Fig.~\ref{figReal}. One can see that the restoration results of KPN+DMN, EDVR+DMN, EDVR*, RviDeNet+DMN and RviDeNet* contain noticeable motion induced artifacts. GCP-Net exhibits less artifact but it fails to reconstruct the textures in the area with high noise level. Benefiting from the two-stage alignment module, our model can more effectively utilize the temporal information and reconstruct more details. We provide more visual results and user study in the supplementary file.

\noindent\textbf{Model Size and Running Speed.}
Table~\ref{table:complex} lists the number of model parameters and running time on 5 frames with UHD resolution by our method and compared methods on a GTX 2080Ti GPU.
EDVR* has the least number of parameters and fastest speed but worst performance. RviDeNet* has the highest computational cost since it consists of a pre-denoising module and non-local modules. Compared with GCP-Net, our model obtains better performance with less training parameters and similar running time. The CA module costs 3.6s and RA+fusion module costs 16.1s. 
 
\begin{table}[!h]
\setlength{\abovecaptionskip}{0.0cm}
\setlength{\belowcaptionskip}{-0.0cm}
\scriptsize
\centering
\caption{Comparison of different CNN models in terms of the number of parameters and running times on 5 frames with input size $2160\times 3840$ and output size $2160\times 3840\times 3$.}
\label{table:complex}
\vspace{0.1cm}
\begin{tabular}{l  c c c c}
\hline
 &EDVR* &RviDeNet* &GCP-Net &Ours \\
\hline
\#. Params &6.28M &57.98M &13.79M &12.05M \\ 
Times(s) &10.4 &124.3 &19.5 &19.7 (3.6+16.1) \\ 
\hline
\end{tabular} 
\vspace{-0.3cm}
\end{table}

\subsection{Ablation Study}
\noindent\textbf{Coarse Alignment vs. Refined Alignment.} To evaluate the role of CA and RA modules in our model, we compare it with three variants, \ie, without using alignment (ours(w/o Align)), only using CA module (ours(w/ \ CA)), and only using RA module (ours(w/ \ RA)). The quantitative comparison on the REDS4 dataset is shown in Table~\ref{table:align_com}.

One can see that ours(w/ \ RA) can achieve great improvement (1.83dB) over ours(w/o Align) when compensating for small motions (\emph{Clip 000}), but achieve limited gain (0.26dB) for sequences with large motion (\emph{Clip 020}). This shows that limited by the small receptive field, DConv fails to effectively use temporal information for scenes with large motion. By using DPBM to search in a large region, ours(w/ \ CA) can more effectively utilize temporal information, and obtain 0.52dB gain than ours(w/ \ RA) for images with large motion (\emph{Clip 020}). However, it achieves smaller improvement (0.31dB) on \emph{Clip 000} which has small motion. We also visualize in Fig.~\ref{disalign} the JDD-B results using models with different alignment modules. One can see that two-stage alignment can reconstruct more textures.

On average, without CA module, ours(w/ \ RA) achieves 1.02dB gain over ours(w/o Align). When CA module is included, ours full model achieves more improvement (1.28dB) than ours(w/ \ CA). Meanwhile, without RA module, ours(w/ \ CA) brings 0.27dB gain than ours(w/o Align), while by considering RA module, our full model brings 0.53dB gain than ours(w/ \ RA). One can see clearly that both alignment steps can improve the JDD-B performance. CA and RA modules are complementary to each other and they can enhance each other when used together.


\vspace{-1mm}
\begin{table}[!h]
\setlength{\abovecaptionskip}{0.0cm}
\setlength{\belowcaptionskip}{-0.0cm}
\footnotesize
\centering
\caption{Comparison of different variants on a clip with small motion (\emph{Clip 000}) and a clip with large motion (\emph{Clip 020}) in the REDS4 dataset, and the average results on the whole REDS4.}
\label{table:align_com}
\begin{tabular}{l | c c c}
\hline
&\emph{Clip 000} &\emph{Clip 020} &Avg\\
\hline
ours(w/o Align) &30.74/0.8662 &32.43/0.8969 &32.88/0.8889 \\
ours(w/ \ CA) &31.05/0.8738 &33.21/0.9080 &33.15/0.8932 \\
ours(w/ \ RA) &32.57/0.9190 &32.69/0.9002 &33.90/0.9077 \\
ours(w/o E2E) &32.67/0.9158 &34.07/0.9217 &34.31/0.9159 \\
\hline
ours(full) &\textbf{32.75/0.9174} &\textbf{34.17/0.9233} &\textbf{34.43/0.9178}\\
\hline
\end{tabular} 
\vspace{-0.1cm}
\end{table}

\begin{figure}[!h]
\setlength{\abovecaptionskip}{0.0cm}
\setlength{\belowcaptionskip}{0.0cm}
\centering
\begin{minipage}[t]{0.18\textwidth} 
\centering
\includegraphics[width=1\textwidth]{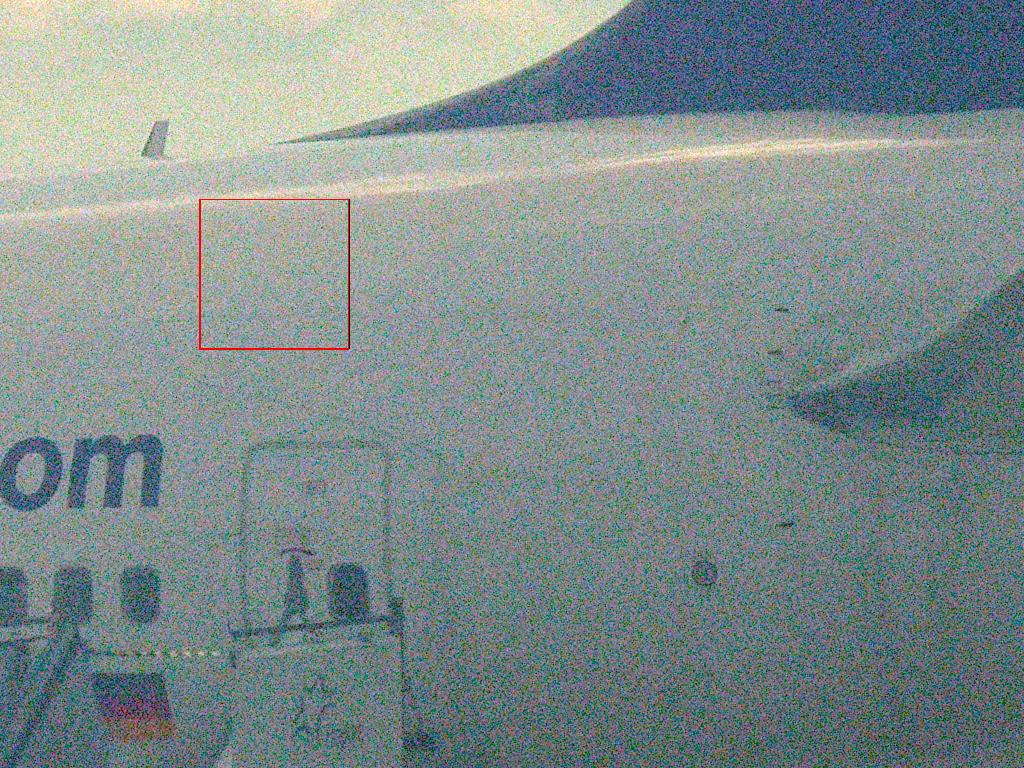}
{\footnotesize  (a) Noisy image}
\end{minipage}\hspace{0.01cm}
\begin{minipage}[t]{0.135\textwidth}
\centering
\includegraphics[width=1\textwidth]{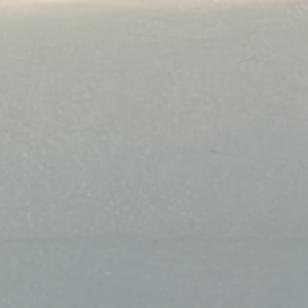}
{\footnotesize  (b) ours(w/o E2E)}
\end{minipage}\hspace{0.01cm}
\begin{minipage}[t]{0.135\textwidth}
\centering
\includegraphics[width=1\textwidth]{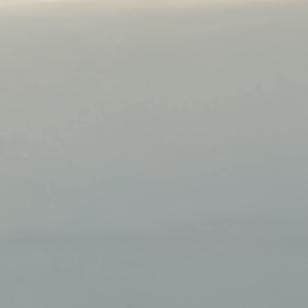}
{\footnotesize  (c) ours}
\end{minipage}\hspace{0.01cm}

\caption{The reconstruction on flat areas with/without end-to-end two-stage alignment learning. Better viewed with zoom-in on screen. Our model relieves the noise interference and obtains cleaner results. }\vspace{-0.3cm}
\label{dise2e}
\end{figure}

\noindent\textbf{End-to-end Learning of Two-stage Alignment.} In order to train our method in an end-to-end manner, we proposed the DPBM module. To evaluate the effectiveness of end-to-end learning, we train a variant, namely ours(w/o E2E), which uses normal BM in the CA module. Since normal BM is non-differentiable, the learnable lightweight downsampling network is replaced by bicubic downsampling. The results are also shown in Table~\ref{table:align_com}. 
One can see that our full model can obtain 0.12dB improvement over ours(w/o E2E), validating that the end-to-end CA module can learn better alignment for restoration. More importantly, since the main difference among flat patches is caused by noise, directly performing BM on noisy patches may align noise and generate artifacts. An example is shown in Fig.~\ref{dise2e}. Owe to the use of a learnable lightweight network for downsampling, our full model can relieve the noise interference especially when aligning flat areas.


\noindent\textbf{Limitation.} When there are objects with large movement in opposite directions in the patch $(P_{t,i})_{i\in I}$, the CA module may fail and our method degrades to use only the RA module in such cases. 

\section{Conclusion}
\label{sec:con}
We presented a differentiable two-stage alignment method for high performance burst image restoration. Due to the limited receptive field, current feature alignment methods cannot sufficiently utilize the temporal information when the burst images have large shift, which is very common for sequences with moving objects and/or 4K resolution. We divided this problem into two relatively easier sub-problems, \ie, coarse and refined alignment, and proposed a two-stage framework for the JDD-B task. In the coarse alignment module, a differentiable progressive block matching module was developed to enlarge the search region while reducing computational cost. Then, a deformable alignment module was developed to deliver pixel-wise alignment. Experiments on both synthetic and real-world burst datasets were conducted. Our method demonstrated clear advantages over existing methods in terms of PSNR/SSIM measures as well as visual quality without increasing much the computational cost. 

\vspace{4mm}
\noindent\textbf{Acknowledgements}
This work is supported by the Hong Kong RGC RIF grant (R5001-18)

{\small

\bibliographystyle{ieee_fullname}
\bibliography{egbib}
}

\end{document}